\pdfoutput=1

\documentclass[11pt,onecolumn]{article}
\usepackage[table,xcdraw]{xcolor}
\usepackage{acl}

\usepackage{times}
\usepackage{latexsym}

\usepackage{paralist,bbding,pifont}
\usepackage{amsthm}
\usepackage{url}
\usepackage{relsize}
\usepackage{amsmath,amsfonts,amssymb}
\usepackage{algorithm,algorithmic}
\usepackage{todonotes}
\usepackage{multirow}
\usepackage{paralist}
\usepackage{graphicx}
\usepackage{etoolbox}
\usepackage{booktabs}
\usepackage{pifont}

\usepackage[normalem]{ulem}
\useunder{\uline}{\ul}{}
\usepackage{mathtools}
\mathtoolsset{showonlyrefs}

\newcommand{\s}{\ensuremath{\mathrm{Interp}}}

\newcommand{\sent}[1]{\ensuremath{s^{(#1)}}}
\newcommand{\ww}[1]{\ensuremath{w^{(#1)}}}

\newcommand{\vv}[1]{\ensuremath{v^{(#1)}}}

\newcommand{\partition}{\mathcal{P}}

\DeclareMathOperator*{\argmin}{arg\,min}
\newtheorem{definition}{Definition}
\newtheorem{example}{Example}
\newcommand{\cmark}{\ding{51}}%
\newcommand{\xmark}{\ding{55}}%
\usepackage[T1]{fontenc}

\usepackage[utf8]{inputenc}

\usepackage{microtype}

\title{Unsupervised Sentence Textual Similarity\\ with Compositional Phrase Semantics}

\author{Zihao Wang \\
  Department of CSE \\
  HKUST\\
  Hong Kong, China\\
  \texttt{zwanggc@cse.ust.hk}
  \And
  Jiaheng Dou \and Yong Zhang~\thanks{\ \ \ Corresponding author.} \\
  BNRist, RIIT, Institute of Internet Industry \\ Department of Computer Science and Technology \\ Tsinghua University, Beijing, China\\
  \texttt{djh19@mails.tsinghua.edu.cn} \\\texttt{zhangyong05@tsinghua.edu.cn}}

\begin{document}
\maketitle
\begin{abstract}
Measuring Sentence Textual Similarity (STS) is a classic task that can be applied to many downstream NLP applications such as text generation and retrieval.
In this paper, we focus on unsupervised STS that works on various domains but only requires minimal data and computational resources.
Theoretically, we propose a light-weighted Expectation-Correction (EC) formulation for STS computation.
EC formulation unifies unsupervised STS approaches including the cosine similarity of Additively Composed (AC) sentence embeddings~\cite{DBLP:conf/iclr/AroraLM17}, Optimal Transport (OT)~\cite{DBLP:conf/icml/KusnerSKW15}, and Tree Kernels (TK)~\cite{le2018acv}.
Moreover, we propose the Recursive Optimal Transport Similarity (ROTS) algorithm to capture the compositional phrase semantics by composing multiple recursive EC formulations. ROTS finishes in linear time and is faster than its predecessors.
ROTS is empirically more effective and scalable than previous approaches.
Extensive experiments on 29 STS tasks under various settings show the clear advantage of ROTS over existing approaches.\footnote{Our code can be found in \url{https://github.com/zihao-wang/rots}.}
Detailed ablation studies demonstrate the effectiveness of our approaches.
\end{abstract}

\section{Introduction}

Sentence Textual Similarity (STS) measures the semantic equivalence between a pair of sentences, which is supposed to be consistent with human evaluation~\cite{DBLP:conf/semeval/AgirreCDG12}.
STS is also an effective sentence-level semantic measure for many downstream tasks such as text generation and retrieval~\cite{DBLP:conf/acl/WietingBGN19,DBLP:conf/emnlp/ZhaoPLGME19,DBLP:conf/aaai/NikolentzosTV20,DBLP:journals/corr/abs-2006-14799}.
In this paper, we focus on unsupervised STS which is expected to compare texts of various domains but only requires minimal data and computational resources.

There are several typical ways to compute unsupervised STS, including 
1) treat each sentence as an embedding by the Additive Composition (\textbf{AC})~\cite{DBLP:conf/iclr/AroraLM17} of word vectors, then estimate the STS of two sentences by their cosine similarity; 
2) treat each sentence as a probabilistic distribution of word vectors, then measure the distance between distributions. 
Notably, Optimal Transport (\textbf{OT})~\cite{peyre2019computational}\footnote{OT-based distance reflects the dissimilarity between sentences and can also be used as STS.} is adopted to compute the STS~\cite{DBLP:conf/icml/KusnerSKW15}. 
OT-based approaches search for the best alignment with respect to the \textbf{word-level semantics} and result in state-of-the-art solution~\cite{yokoi2020word}.

In this paper, we argue that \textbf{phrase-level semantics} should also be exploited to fully understand the sentences.
For example, ``optimal transport'' should be considered as a mathematical term rather than two independent words.
Specifically, the phrase chunk is composed of lower-level chunks and is usually represented as a node in tree structures. 
The aforementioned AC and OT-based STS methods are too shallow to include such structures. 
Tree Kernels (\textbf{TK})~\cite{le2018acv} consider the parsed \emph{syntax labels}.
However, it boils down to syntax-based but sub-optimal word alignment under our comparison experiment.

Recent advancement of Pretrained Language Models (PLMs) also demonstrate the importance of contextualization~\cite{DBLP:conf/naacl/PetersNIGCLZ18,DBLP:conf/naacl/DevlinCLT19,DBLP:conf/emnlp/Ethayarajh19}. 
PLMs can be further adopted to STS tasks by supervised fine-tuning~\cite{DBLP:conf/naacl/DevlinCLT19}, under carefully designed transfer learning~\cite{reimers2019sentence} or domain-adaptation~\cite{li2020sentence,gao2021simcse}. 
Without those treatments, the performances of PLM-based STSs are observed to be very poor~\cite{yokoi2020word}. 
Meanwhile, PLM-based STSs suffer from high computational costs to fit large amounts of high-quality data, which might prevent them from broader downstream scenarios.

In this paper, we propose a set of concepts and similarities to exploit the phrase semantics in the unsupervised setup.
Our contributions are four folds:
\begin{compactdesc}
\item[Unified formulation] We unify three types of unsupervised STS models (AC~\cite{DBLP:conf/iclr/AroraLM17}, OT~\cite{yokoi2020word} and TK~\cite{le2018acv}) by the EC similarity in Section~\ref{sec:ec-unification}. 
EC similarity uncovers the strengths and weaknesses of the three approaches.

\item[Phrase vectors and their alignment] We generalize the idea of word alignment to phrase alignment in Section~\ref{sec:word2phrase}. 
After the formal definition of Recursive Phrase Partition (RPP), we compose the phrase weights and vectors by those from finer-grained partitions under the \textit{invariant additive phrase composition} and generalize the word alignment to phrase alignment. 
Empirical observations show that EC similarity is an effective formulation to interpolate the existing unsupervised STS, and yields better performances.

\item[Recursive Optimal Transport] We propose the Recursive Optimal Transport Similarity (ROTS) in Section~\ref{sec:rotsts} based on the phrase alignment introduced in Section~\ref{sec:word2phrase}. 
ROTS computes the EC similarity at each phrase partition level and ensembles them. 
Notably, Prior Optimal Transport (Prior OT) is adopted to guide the finer-grained phrase alignment by the coarser-grained phrase alignment at each expectation step of EC similarity.

\item[Extensive experiments] We show the comprehensive performance of ROTS on a wide spectrum of experimental settings in Section~\ref{sec:exp} and the Appendix, including 29 STS tasks, five types of word vectors, and three typical preprocessing setups.
Specifically, ROTS is shown to be better than all other unsupervised approaches including $\mathrm{BERT}$ based STS in terms of both effectiveness and efficiency.
Detailed ablation studies also show that our constructive definitions are sufficiently important and the hyper-parameters can be easily chosen to obtain the new SOTA performances.
\end{compactdesc}

\section{Related Work}


Embedding the symbolic words into continuous space to present their semantics~\cite{DBLP:conf/nips/MikolovSCCD13,DBLP:conf/emnlp/PenningtonSM14,bojanowski2017enriching} is one of the breakthroughs of modern NLP.
Notably, it shows that the vector (or semantics) of a phrase can be approximated by the \emph{additive composition} of the vectors of its containing words~\cite{DBLP:conf/nips/MikolovSCCD13}.
Thus, word embeddings can be further utilized to describe the semantics of texts beyond the word level. Several strategies were proposed to provide sentence embeddings.

\noindent\textbf{Additive Composition.} Additive composition of word vectors~\cite{DBLP:conf/iclr/AroraLM17} forms effective sentence embeddings. 
The cosine similarity between the sentence embeddings has been shown to be a stronger STS under transferred\cite{DBLP:conf/iclr/WietingBGL15a,DBLP:conf/acl/GimpelW18} and unsupervised settings~\cite{DBLP:conf/iclr/AroraLM17,DBLP:conf/rep4nlp/EthayarajhH18} than most of the deep learning approaches~\cite{DBLP:conf/emnlp/SocherPWCMNP13,DBLP:conf/icml/LeM14,DBLP:conf/nips/KirosZSZUTF15,DBLP:conf/acl/TaiSM15}.

\noindent\textbf{Optimal Transport.} By considering sentences as distributions of embeddings, the similarity between sentence pairs is the consequence of optimal transport of sentence distributions~\cite{DBLP:conf/icml/KusnerSKW15,DBLP:conf/nips/HuangGKSSW16,DBLP:conf/emnlp/WuYXXBCRW18,yokoi2020word}. 
OT models find the optimal alignment with respect to word semantics via their embeddings and have the SOTA performances~\cite{yokoi2020word}.

\noindent\textbf{Syntax Information.}
One possible way to integrate contextual information in a sentence is to explicitly employ syntactic information. 
Recurrent neural networks~\cite{DBLP:conf/emnlp/SocherPWCMNP13} were proposed to exploit the tree structures in the supervised setting but were sub-optimal than AC-based STS.
Meanwhile, tree kernels~\cite{moschitti2006efficient,croce2011structured} can measure the similarity between parsing trees. 
Most recently, ACV-tree kernels~\cite{le2018acv} combine word embedding similarities with parsed constituency labels. 
However, tree kernels compare all the sub-trees and suffer from high computational complexity.

\noindent\textbf{Pretrained Language Models}
This paradigm produces contextualized sentence embeddings by aggregating the word embeddings repeatedly with the deep neural networks~\cite{vaswani2017attention} trained on large corpuses~\cite{DBLP:conf/naacl/DevlinCLT19}.
In the unsupervised setting, PLMs are sub-optimal compared to SOTA OT-based models~\cite{yokoi2020word}.
One of the common strategies to improve the performance is to adjust PLM-generated embedding according to a large amount of external data such as transfer learning~\cite{reimers2019sentence}, flow~\cite{li2020sentence}, whitening~\cite{su2021whitening}, and contrastive learning~\cite{gao2021simcse}. 
However, this domain adaptation paradigm requires a complex training process and the performance is highly affected by the similarity between the target test data and external data~\cite{li2020sentence,gao2021simcse}. 

\section{Unification of Unsupervised STS Methods}\label{sec:ec-unification}

\begin{figure*}[t]
	\centering
	\includegraphics[width=0.9\linewidth]{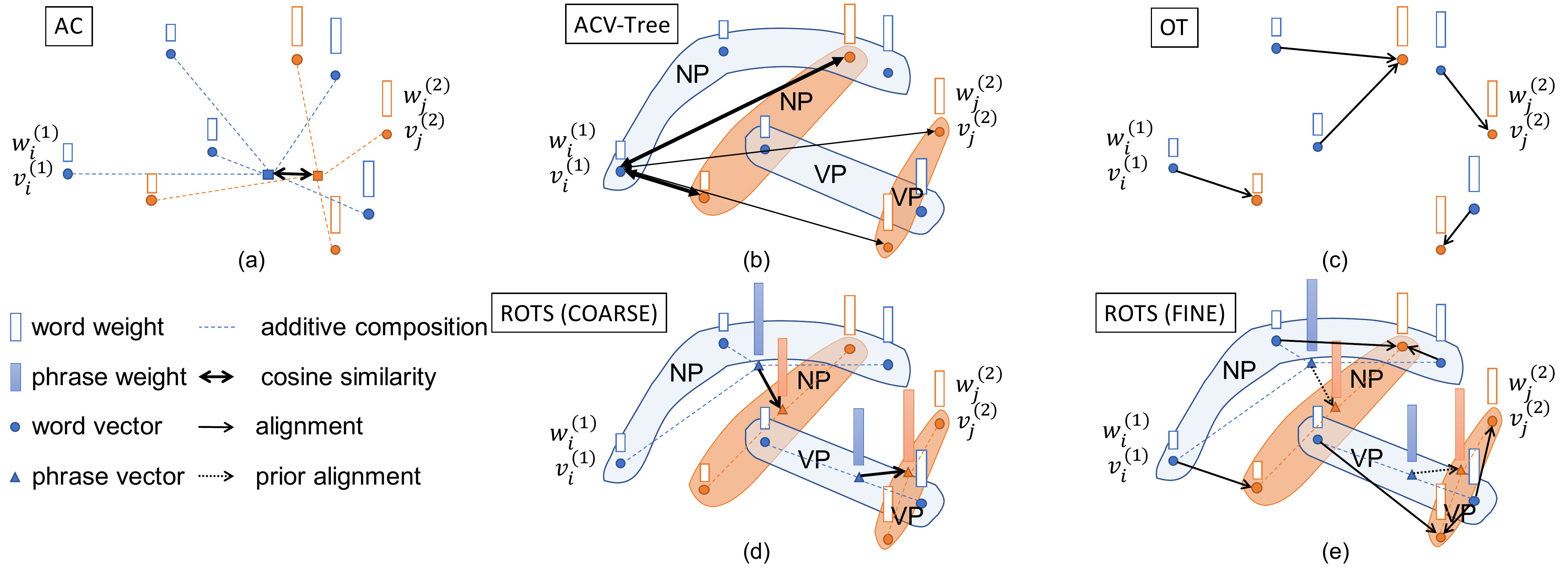}
	\caption{Different unsupervised STS methods with blue elements for $ \sent{1} $ and orange elements for $ \sent{2} $. 
	\textbf{(a)} AC~\cite{DBLP:conf/iclr/AroraLM17}: cosine similarity between additively composed sentence embeddings. 
	\textbf{(b)} ACV-Tree~\cite{le2018acv}: weighted averaging pairwise word similarity.
	Similarities from $\vv{1}_i$ to vectors in $ \sent{2} $ are shown. 
	More weights are assigned to pairs contained in the same constituency structure, indicated by thicker arrows. 
	\textbf{(c)} OT~\cite{yokoi2020word}: compute the optimal transport alignment of words by solving problem~\eqref{eq:ot}. 
	\textbf{(d)} ROTS at coarser hierarchy: the OT alignment of phrases vectors and weights. 
	\textbf{(e)} ROTS at finer hierarchy: fine-level OT alignment based on the prior of coarse-level alignment in (d).}\label{fig:illustrate}
\end{figure*}

Given a pair of sentences $(\sent{1},\sent{2})$, we are expected to estimate their similarity score $s\in [0, 1]$. 
For sentence $\sent{1}$ (or $\sent{2}$), we have vector $\{\vv{1}_i\}_{i=1}^m$ (or $\{\vv{2}_j\}_{j=1}^n$) and weight $ \{\ww{1}_i\}_{i=1}^m$ (or $\{\ww{2}_j\}_{j=1}^n$).
We quickly review three types of unsupervised STS in Section~\ref{sec:review} (see Figure~\ref{fig:illustrate} (a-c)), then unify them by the Expectation-Correction similarity in Section~\ref{sec:ec-similarity}.

\subsection{Review of Three Types of STS}\label{sec:review}
\noindent\textbf{Additive Composition (AC)} 
AC methods~\cite{DBLP:conf/iclr/AroraLM17,DBLP:conf/rep4nlp/EthayarajhH18} firstly compute the sentence embedding $ x^{(\cdot)} = \sum_i \ww{\cdot}_i \vv{\cdot}_i $, then estimate the similarity by the cosine similarity $ s_{AC} = \cos(x^{(1)}, x^{(2)})$, see Figure~\ref{fig:illustrate} (a). 

\noindent\textbf{Optimal Transport (OT)}
Given pairwise word distance matrix $ D = D_{ij} $ and two marginal distributions $ \mu_i $ and $ \nu_i $, the optimal transport alignment $ \Gamma_{OT} $ is computed by solving the following minimization problem~\cite{DBLP:conf/icml/KusnerSKW15}.
\begin{gather}\label{eq:ot}
\Gamma_{OT} = \argmin_{\Gamma_{ij} \geq 0} \sum_{ij} \Gamma_{ij} D_{ij}, \\
\text{s.t. } \sum_j \Gamma_{ij} = \mu_i, \sum_i \Gamma_{ij} = \nu_j.
\end{gather}
The higher $\Gamma_{OT,ij}$ means that the alignment from $i$-th word in $ \sent{1} $ to $j$-th word in $ \sent{2} $ is preferred, because those two words are semantically closer, see Figure~\ref{fig:illustrate} (c).
Different choices of $ D, \mu, \nu $ lead to different distances. 
The SOTA OT-based STS is the Word Rotator's Distance (WRD)\footnote{Without further specification, OT is referred to WRD}~\cite{yokoi2020word}, which solves Problem~\eqref{eq:ot} with $ D_{ij}  = 1-\cos(\ww{1}_i, \ww{2}_j) $ and
\begin{gather}\label{eq:wrd}
 \mu_i  = \frac{\ww{1}_i \|\vv{1}_i\|_2}{\sum_k \ww{1}_k \|\vv{1}_k\|_2},  \\ \nonumber
 \nu_j  = \frac{\ww{2}_j \|\vv{2}_j\|_2}{\sum_k \ww{2}_k \|\vv{2}_k\|_2}.
\end{gather}
The similarity is 
\begin{align}\label{eq:sim:ot}
	s_{OT} = \sum_{ij} \Gamma_{OT,ij} \cos(\ww{1}_i, \ww{2}_j).
\end{align}
WRD is equivalent to AC if and only if each sentence contains one word~\cite{yokoi2020word}.

\noindent\textbf{Tree Kernel (TK)}
General tree kernels compare the syntactic parsing information~\cite{moschitti2006efficient,croce2011structured}. 
Recently, ACV-Tree~\cite{le2018acv} combines word-level semantics with syntax information by a simplified partial tree kernel~\cite{moschitti2006efficient}, see Figure~\ref{fig:illustrate} (b). 
Word similarities from the same structure, i.e. NP, are repeatedly counted and thus more important. 
Then the similarity score can be re-written as 
\begin{align}\label{eq:sim:tk}
	s_{TK} = \sum_{ij} \Gamma_{TK, ij} \cos(\ww{1}_i, \ww{2}_j)
\end{align}
where $ \Gamma_{TK} $ is the normalized weight matrix generated by the tree kernel~\footnote{In this paper, TK indicates the ACV-Tree kernel}.

\subsection{Expectation Correction (EC)}~\label{sec:ec-similarity}
Three approaches discussed above, though motivated in different ways, can be seen as a linear aggregation of pair-wise cosine similarities of words. We unified them into the following EC similarity with two steps called \textbf{expectation} and \textbf{correction}.

\begin{table*}[t]
	\centering
	\caption{The comparison of different approaches. 
	}\label{tb:diff-sts}
	\scriptsize
\begin{tabular}{|l|lll|l|l|}
	\hline
	\multirow{2}{*}{Method} & \multicolumn{3}{c|}{Inter-sentence Expectation} & \multirow{2}{*}{Intra-sentence Correction} & \multirow{2}{*}{Tiime Complexity} \\\cline{2-4}
	& Word Semantics   & Phrase Semantics  & Syntax  &                                            &                            \\\hline
	AC~\cite{DBLP:conf/iclr/AroraLM17,DBLP:conf/rep4nlp/EthayarajhH18} & \xmark & \xmark & \xmark & \cmark & $ O(m+n) $\\
	OT~\cite{DBLP:conf/icml/KusnerSKW15,yokoi2020word} & \cmark & \xmark & \xmark & \xmark & $ O(mn) $\\
	TK~\cite{le2018acv} & \xmark & \xmark & \cmark & \xmark & $ O(mn) $\\
	ROTS (ours) & \cmark & \cmark & \cmark & \cmark & $ O(m+n) $\\\hline
\end{tabular}
\end{table*}

\noindent\textbf{Expectation} Both ACV-Tree (see Equation~\eqref{eq:sim:tk}) and OT (see Equation~\eqref{eq:sim:ot}) aggregate pairwise word similarities by the alignment matrix $\Gamma_{TK}$ and $\Gamma_{OT}$.
AC also implies the implicit word alignment $\Gamma_{AC}$, the cosine similarity can be further decomposed by plugging in the sentence vectors:

\begin{align}
    \cos(x^{(1)}, x^{(2)}) &= \frac{\langle \sum_i \ww{1}_i \vv{1}_i, \sum_j \ww{2}_j \vv{2}_j \rangle}{\lVert x^{(1)} \rVert \lVert x^{(2)} \rVert} \\
    &= C\sum_{ij} \Gamma_{AC,ij} \cos(\vv{1}_i, \vv{2}_j)~\label{eq:cosot}
\end{align}
where $\Gamma_{AC,ij} = \mu_i  \nu_j $, $\mu$ and $\nu$ are defined in Equation~\eqref{eq:wrd}.
This observation connects AC to the \emph{expectation} of word similarities~\footnote{Equation~\eqref{eq:cosot} motivates the marginal conditions of WRD in a different way}.
Hence, the key of \textbf{expectation} step, is to compute \textbf{inter-sentence} word alignment matrix $ \Gamma $. 
Specifically, $ \Gamma_{AC} $ is implicitly induced by weights and vector norms without considering the semantics or syntax between words, $ \Gamma_{TK} $ is constructed by comparing node labels in syntax trees, and $ \Gamma_{OT} $ is obtained by optimizing word semantics. (See Table~\ref{tb:diff-sts})

\noindent\textbf{Correction}
In Equation~\eqref{eq:cosot}, the coefficient 
\begin{align}
    C = \frac{\sum_k \ww{2}_k\lVert\vv{2}_k\rVert}{\lVert\sum_k \ww{2}_k\vv{2}_k\rVert} \frac{\sum_k \ww{1}_k\lVert\vv{1}_k\rVert}{\lVert\sum_k\ww{1}_k\vv{1}_k\rVert} = \sqrt{K_1 K_2}
\end{align}
also has special interpretation.
For the specific sentence $i=1,2$, the coefficient $K_i$ can be rewritten as
\begin{align}
&K_{i} - 1 = \frac{(\sum_k \ww{i}_k\lVert\vv{i}_k\rVert)^2}{\lVert\sum_k \ww{i}_k\vv{i}_k\rVert^2} - 1 \\
&= \sum_{k\neq m} \frac{\ww{i}_k \ww{i}_m \lVert \vv{i}_k\rVert \lVert \vv{i}_m\rVert}{\lVert\sum_k \ww{i}_k\vv{i}_k\rVert^2} \left[1 - \cos(\vv{i}_k, \vv{i}_m)\right].\label{eq:k}
\end{align}
We have $K_i \geq 1$ and the equality holds if and only if all word vectors are in the same direction, i.e. they are semantically close. 
$K_i$ increases as the semantics of words in a sentence become more diverse. 
In the latter situation, the sentence similarity tends to be underestimated since unnecessary alignments are forced by the joint distribution. 
The coefficient $C$ corrects this \textbf{intra-sentence} semantics. 
This correction step distinguishes AC from OT and TK approaches (see Table~\ref{tb:diff-sts}).

Then we introduce the EC similarity by combining E-step and C-step as follows:

\begin{definition}[EC similarity]~\label{def:ecformulation}
The EC similarity of STS is defined by:
\begin{align}
    \tilde C \sum_{ij} \Gamma_{ij} \cos(\vv{1}_i, \vv{2}_j),~\label{eq:ec}
\end{align}
where $\Gamma$ is the word alignment matrix for the \textbf{expectation} and $\tilde C = (\alpha C + 1 - \alpha)$ is the coefficient for \textbf{correction}, hyper-parameter $\alpha \in [0, 1]$ linearly interpolates the $C$ and $1$ and controls the strength of correction.
\end{definition}

\section{From Word to Phrase Alignment}\label{sec:word2phrase}


In this section, we extend the word alignment to the phrase alignment. 
We define the phrase partitions of sentences with the recursive structure from any tree.
Then we define the phrase weight and vector by the additive composition of sub-phrase (or word) weights and vectors.


\subsection{Recursive Phrase Partitions (RPP)}\label{sec:partition}

\begin{figure*}[t]
\centering
\includegraphics[width=0.85\linewidth]{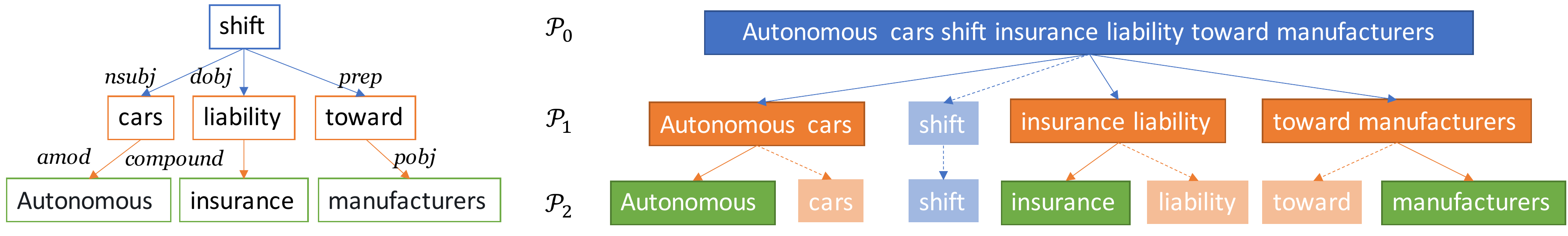}
\caption{Dependency tree (left) by SpaCy~\cite{spacy2} and recursive phrase partitions (right)}\label{fig:tree-partition}
\end{figure*}

For sentence $s=[t_1, ..., t_n]$ containing $n$ tokens $t_i, 1\leq i\leq n$, we define the Recursive Phrase Partitions (RPP) as a set of partitions $\{\partition_0, \partition_1, ..., \partition_{L}\}$ of the sentence $s$, where $ \partition_l $ is the partition at $l$-th level, $ 1\leq l\leq L $. 
Specifically, $\partition_l = [P_{l,1}, ..., P_{l,q}]$ contains a sequence of phrases, where the $q$-th phrase $ P_{l,q} = s[b_{l,q}: e_{l,q}] $ is the span in $ s $ from the beginning index $ b_{l,q} $ to the ending index $ e_{l,q} $.
So we have two properties:
\begin{compactenum}
\item Concatenating all phrases recovers the sentence, that is $\oplus_q P_{l,q} = s$, where $ \oplus $ is the string concatenation.
\item For two different levels, i.e. $0 \leq l_1 < l_2$,\footnote{We denote the root is level $0$. The level index increases as the tree goes deeper.} any phrase in level $l_2$ is contained in the unique phrase in level $l_1$.
\end{compactenum}
In our definition, the $ \partition_0 = [s] $ and $ \partition_{L} = [t_1, .., t_n] $ are the coarsest partition and the finest partition, respectively.
The second property guarantees that the recursive phrase partitions can be nested so that each phrase can be recursively divided.
RPP can be constructed from any tree representation of the sentence, including constituency tree, dependency tree, or even naive binary separation of token sequences.
Figure~\ref{fig:tree-partition} shows an example of RPP from a dependency tree. 
Some phrases (such as `shift' in $\partition_2$) are added to satisfy the first property.

\subsection{Compositional Phrase Semantics}\label{sec:compose-phrase}
Once the RPP structure of a sentence is given, we define the vector $ \tilde{v} $ and weight $ \tilde{w} $ for each phrase. Our definition is invariant with respect to the AC sentence embedding, that is, AC sentence embedding $ x $ is invariant to the phrase partition $ \partition_l $ of the sentence.
\begin{align}\label{eq:inv-phrase}
x & = \sum_i^n w_i v_i 
= \sum_q \tilde{w}_{l,q} \tilde{v}_{l,q},
\end{align}
where the phrase weights and vectors are given by
\begin{align}
	\tilde{w}_{l,q} = \sum_{i= b_{l,q}}^{e_{l,q}} w_i, \tilde{v}_{l,q} = \sum_{i = b_{l,q}}^{e_{l,q}} w_i v_i/\tilde{w}_{l,q}.
\end{align}
In this way, the sentence vector can also be represented by the additive composition of phrase vectors and weights, where each phrase vector can be again composed by the word vectors additively.
Our definitions of phrase weights and vectors recursively aggregate the information from finer-grained level (i.e. `autonomous' and `cars') information to coarser-grained level (i.e. `autonomous cars').
Furthermore, our discussion about EC similarity in Section~\ref{sec:ec-similarity} at the word level can also be generalized to any phrase partitions. 
That is, we can use the EC similarity to consider the inter-sentence \textit{phrase} alignment and then correct the intra-sentence \textit{phrase} semantics of each partition.

\section{Recursive Optimal Transport and STS}\label{sec:rotsts}
In this section, we connect the dots by applying EC similarity in Section~\ref{sec:ec-similarity} to phrase alignment in Section~\ref{sec:word2phrase} on tree structures. Specifically, we present Recursive Optimal Transport Similarity (ROTS) which computes the phrase alignment at each $(l+1)$-th level phrase partition with the guidance of the phrase alignment at the $l$-th level.

\subsection{Prior Optimal Transport (Prior OT)}~\label{sec:prior-ot}
Prior OT~\cite{zhao2020semi} was firstly proposed to pass prior information when minimizing the entropy-regularized Wasserstein loss.
When it comes to the OT-based STS, we re-consider the objective function in Problem~\eqref{eq:ot} with an additional prior alignment $\Pi$:
\begin{align}~\label{eq:wrd-pot}
\sum_{ij} \Gamma_{ij} D_{ij} + \epsilon \mathrm{KL}(\Gamma \| \Pi),
\end{align}
where $\mathrm{KL}(\Gamma\| \Pi) = - \sum_{ij} \Gamma_{ij} \log{\Pi_{ij}} - H(\Gamma)$ is the KL-divergence between the phrase alignment $ \Gamma $ and the prior alignment $ \Pi $, and $H(\cdot)$ is the entropy.
$\epsilon$ is the hyper-parameter that controls how close the obtained $ \Gamma^* $ is to $ \Pi $. 
When $\epsilon=0$, Equation~\eqref{eq:wrd-pot} falls back to Equation~\eqref{eq:ot}, and when $ \epsilon $ is sufficiently large, the optimal $\Gamma^*$ is sufficiently close to $\Pi$ in terms of KL-divergence.

Notably, the objective in Equation~\eqref{eq:wrd-pot} can be minimized by the Sinkhorn algorithm~\cite{DBLP:conf/nips/Cuturi13,zhao2020semi}. 
Compared to tree kernels~\cite{moschitti2006efficient,croce2011structured,le2018acv}, Sinkhorn algorithm is based on matrix operations such that it can be accelerated by GPUs~\cite{DBLP:conf/nips/Cuturi13}.
Sinkhorn algorithm has time complexity $O(mn / \epsilon^{2})$~\cite{DBLP:conf/icml/DvurechenskyGK18}. 
In our practice, we usually choose the large prior strength, i.e. $\epsilon > 1$ that allows faster convergence.

We can interpolate WRD and AC with the help of Prior OT under EC similarity.
\begin{example}[EC Interpolation of WRD and AC]~\label{eg:ec-interp}
    Given a prior matrix $\Pi = \Gamma_{AC}$, we first compute the alignment $\Gamma_\epsilon$ by minimizing Equation~\eqref{eq:wrd-pot} with WRD's choice of $D, \mu, \nu$ in Equation~\eqref{eq:wrd}. Then we compute the EC interpolation similarity by 
\begin{align}
\s = \tilde C \sum_{ij} \Gamma_{\epsilon, ij} \cos(\vv{1}_i, \vv{2}_j),
\end{align}
where $\epsilon>0$ is the prior strength in Equation~\eqref{eq:wrd-pot}. 
When $(\alpha,\epsilon) = (0, 0)$, $ 1-\s =s_{OT} $ \footnote{$\s$ itself also leads to the identical STS evaluation as $ s_{OT} $ in terms of correlation.}. 
When $(\alpha,\epsilon) = (1, +\infty)$ , $\s=s_{AC}$.
\end{example}

\subsection{Recursive Optimal Transport Similarity}\label{sec:rots}
Given two sentences $\sent{1}, \sent{2}$ with their RPPs $\{\partition^{(1)}_0, \partition^{(1)}_1, ..., \partition^{(1)}_{L_1}\}$ and $\{\partition^{(2)}_0, \partition^{(2)}_1, ..., \partition^{(2)}_{L_2}\}$, 
ROTS considers partition pairs $(\partition^{(1)}_k, \partition^{(2)}_k)$ from the coarsest $k = 0$ level to the finest $ k = d \leq \min(L_1, L_2)$ level, where $ d $ is a hyper-parameter.
Given the computed $ k $-th alignment matrix $\Gamma^{(k)}$ of $(\partition^{(1)}_k, \partition^{(2)}_k)$, ROTS constructs the following prior alignment $\Pi^{(k+1)}$ for next EC computation $(\partition^{(1)}_{k+1}, \partition^{(2)}_{k+1})$.
\begin{align}
\Pi^{(k+1)}_{m_i n_j} = \frac{\mu_{k+1, m_i}\nu_{k+1, n_j} \Gamma^{(k)}_{ij} }{\sum_{ \tilde m_i \in P^{(1)}_{k, i}, \tilde n_j \in P^{(2)}_{k, j}} \mu_{k+1, \tilde m_i}\nu_{k+1, \tilde n_j}}.\label{eq:coarse-to-fine}
\end{align}
Specifically, the $ (i,j) $ phrase alignment score $ \Gamma_{ij}^{(k)} $ at $ k $-th level will be separated to the sub-phrase alignment $ (m_i, n_j) $ at the $ (k+1) $-th level according to the marginal $ \mu_{k+1, m_i} $ and $ \nu_{k+1, n_j} $, where $ m_i,n_j $ are the index of the sub-phrase of $ i, j $ respectively.
With the coarse-to-fine prior $\Pi^{(k+1)}$, ROTS computes the phrase alignment matrix $\Gamma^{(k+1)}$ at the $(k+1)$-th level by Prior OT (Equation~\eqref{eq:wrd-pot}). 
The computation process of ROTS is shown in Algorithm \ref{algo:rots}.
For $k=0$, each sentence has a single vector, the alignment matrix $ \Gamma^{(0)} = 1 $ is a $ 1\times 1 $ matrix. 
The complexity of ROTS is $ O(m+n+d(\rho^d/\epsilon)^2) $ where $ \rho $ is the maximum branching number of the tree and is usually small for natural language. 
When the hyper-parameter $ d $ is fixed, the complexity of Algorithm~\ref{algo:rots} grows linearly with the sentence length $ m $ and $ n $ (see Table~\ref{tb:diff-sts}).

Our ROTS is featured by finding the finer-level phrase alignment under the guidance of the coarser-level phrase alignment. 
Unlike the tree kernels~\cite{le2018acv} that highly rely on syntax trees and syntax labels, ROTS is based on the EC phrase alignment at different phrase partition levels that are induced by a syntax tree. 
Specifically, the phrase alignments are obtained from the phrase semantic information, i.e. weights and vectors rather than plain syntax labels (see Table~\ref{tb:diff-sts}). 

\begin{algorithm}[t]
\caption{Recursive OT Similarity}\label{algo:rots}
    \begin{algorithmic}[1]
        \REQUIRE Two sentences $\sent{1}, \sent{2}$ with recursive phrase partitions $\{\partition^{(1)}_0, \partition^{(1)}_1, ...\}$ and $\{\partition^{(2)}_0, \partition^{(2)}_1, ... \}$, depth $d$ and prior strengths $\epsilon_k, k=1, ..., d$, correction strength $\alpha$.
        \ENSURE $\mathrm{ROTS}_k$ at each level $k$.
        \STATE Prepare the weights and vectors at level $0$.
        \STATE Initialize 0-th level alignment $\Gamma^{(0)} \gets 1$.
        \FOR{$k \gets 1,...,d$}
        \STATE Prepare the weights and vectors at level $k$.
        \STATE Get $k$-th prior $\Pi^{(k)}$ by Eq.~\eqref{eq:coarse-to-fine} from $\Gamma^{(k-1)}$.
        \STATE Get $k$-th alignment $\Gamma^{(k)}$ by Eq.~\eqref{eq:wrd-pot} with $\epsilon_k$.
        \STATE Get $\mathrm{ROTS}_k$ by Eq.~\eqref{eq:ec} with $\tilde C = \alpha C + 1 - \alpha$, where $C = \frac{\sum_k \ww{2}_k\lVert\vv{2}_k\rVert}{\lVert\sum_k \ww{2}_k\vv{2}_k\rVert} \frac{\sum_k \ww{1}_k\lVert\vv{1}_k\rVert}{\lVert\sum_k\ww{1}_k\vv{1}_k\rVert}$.
        \ENDFOR
    \end{algorithmic}
\end{algorithm}

\section{Experiments}\label{sec:exp}

We first present the experimental setting of unsupervised STS. Then we conduct the benchmark study of all unsupervised STS approaches. Detailed ablation studies justify the effect of ROTS. In the appendix, further discussions on the impact of word vectors, and preprocessing steps are included. 

\subsection{Experimental Settings}~\label{sec:exp-setting}

\noindent\textbf{Text processing} SpaCy~\cite{spacy2} is a open-source text processing toolkit including rich functionality such as tokenization and dependency parsing. It is very suitable for preprocessing pipelines. The text processing model in \texttt{en\_core\_web\_sm} is used.

\noindent\textbf{Word vectors}
Word2Vec~\cite{DBLP:conf/nips/MikolovSCCD13}, GloVe~\cite{DBLP:conf/emnlp/PenningtonSM14}, and fastText~\cite{bojanowski2017enriching} are considered in the unsupervised STS cases.
Two word vectors trained on transferred learning settings, i.e. PSL~\cite{DBLP:journals/tacl/WietingBGL15} and ParaNMT~\cite{DBLP:conf/acl/GimpelW18}, are considered in the transferred STS cases.
Further information can be found in Appendix~\ref{app:wv}.

\noindent\textbf{Preprocessing}
The scope of our pre-processing steps extends the ``vector converters'' in \cite{yokoi2020word}. Those preprocessing steps can all be applied to EC similarity and are detailed in Appendix~\ref{app:preprocessing}.
Three typical setups are selected, including SUP~\cite{DBLP:conf/rep4nlp/EthayarajhH18}, SWC~\cite{yokoi2020word} and WR~\cite{DBLP:conf/iclr/AroraLM17}.

\noindent\textbf{Datasets} We consider (1) \textbf{STSB} dev and test set in STS-Benchmark~\cite{DBLP:journals/corr/abs-1708-00055}; (2) \textbf{STS[year]} STS from 2012 to 2016~\cite{DBLP:conf/semeval/AgirreCDG12,DBLP:conf/starsem/AgirreCDGG13,DBLP:conf/semeval/AgirreBCCDGGMRW14,DBLP:conf/semeval/AgirreBCCDGGLMM15,DBLP:conf/semeval/AgirreBCDGMRW16}; (3) \textbf{SICK}~\cite{DBLP:conf/semeval/MarelliBBBMZ14}; (4) \textbf{Twitter}~\cite{DBLP:conf/semeval/XuCD15}. 
Details can be found in Appendix~\ref{app:dataset}.
Each dataset includes several sub-tasks, and there are 29 tasks in total.

\noindent\textbf{Related baselines} Some unsupervised STS baselines are closely related to EC similarity, including COS (SIF~\cite{DBLP:conf/iclr/AroraLM17}, uSIF~\cite{DBLP:conf/rep4nlp/EthayarajhH18}),  ACV-Tree~\cite{le2018acv}, and WRD~\cite{yokoi2020word}. WMD~\cite{DBLP:conf/icml/KusnerSKW15} is important but not included since WMD has been shown clearly suboptimal to WRD~\cite{yokoi2020word}.

\noindent\textbf{Other Unsupervised Baselines} BERT's final-layer and last-2-layers embeddings (BERT and BERT-last2ave)~\cite{li2020sentence} BERTScore~\cite{DBLP:conf/iclr/ZhangKWWA20}, DynaMax-Jaccard~\cite{DBLP:conf/iclr/ZhelezniakSSMFH19}, Center Kernel Alignment (CKA)~\cite{DBLP:conf/emnlp/ZhelezniakSBSH19} and Kraskov-St\"ogbauer–Grassberger~\cite{kraskov2004estimating} (KSG) cross entropy estimation~\cite{zhelezniak2020estimating}.


\noindent\textbf{Default hyper-parameters}
We summarize the result with different parameters. Results show that excellent scores are achieved with $\alpha=1$, $d=4$ and $\epsilon_k = 10, 1\leq k\leq L$.

\subsection{Unsupervised Benchmark}\label{exp:up-benchmark}
An unsupervised STS benchmark study is conducted over STSB, SICK, and STS by years (STS12-16).
Twitter is not included since most of the baselines did not report the score. fastText is chosen as the pretrained word vector.

\begin{table*}[t]
	\centering
	\caption{Pearson's $r \times 100$ for ROTS and related unsupervised baselines. Best cases are in boldface.}~\label{tb:benchmark-related}
	\small
	\begin{tabular}{|l|ll|lllll|}
		\hline
		Similarity & STSB & SICK  & STS12  & STS13  & STS14  & STS15  & STS16  \\
		\hline
		ACV-Tree~\cite{le2018acv} & - & - & 61.60 & - & 72.83 & 75.80 & - \\
		BERTScore fastText~\cite{DBLP:conf/iclr/ZhangKWWA20} & 53.86 & 64.69 & 51.95 & 45.86 & 61.66 & 69.00 & - \\
		\hline
		SIF\cite{DBLP:conf/iclr/AroraLM17} & 70.13 & {73.20} & 63.46 & 59.30 & 72.95 & 73.27 & 70.79 \\
		uSIF\cite{DBLP:conf/rep4nlp/EthayarajhH18} & {73.47}  & 72.73 & 63.24 & 61.41 & {74.37}  & 76.33 & 73.47 \\\hline
		WRD+SWC\cite{yokoi2020word} & 74.58 & 67.09 & 63.80 & 57.55 & 71.06 & 77.65 &   75.46 \\
		WRD+SUP\cite{yokoi2020word} & 74.80 & 67.67 & \textbf{64.03} & 58.50 & 71.32 & 77.65 &  75.38 \\
		WRD+WR\cite{yokoi2020word}  &73.13 & 68.73 & 63.81 & 58.09 & 70.60 & 77.28 & 74.48 \\\hline
		ROTS+SWC+mean & \textbf{{75.33}} & 71.79  & 63.91 & \textbf{62.29} & {74.30} & \textbf{{77.96}} & \textbf{75.95} \\
		ROTS+SUP+mean  & 74.25  & 73.13  & 63.52  & 61.49 & \textbf{{74.44}} &  76.75  & 74.28 \\
		ROTS+WR+mean & 71.52 & \textbf{{73.84}} & 63.77 & 59.58 & 73.15 & 73.91 & 71.97 \\\hline
	\end{tabular}
\end{table*}

\begin{table*}[t]
	\centering
	\caption{Spearman's $\rho\times 100$ for ROTS and other unsupervised baselines. Best cases are in boldface.}~\label{tb:benchmark-others}
	\small
	\begin{tabular}{|l|ll|lllll|}
		\hline
		Similarity & STSB & SICK  & STS12  & STS13  & STS14  & STS15  & STS16  \\ \hline
		$\mathrm{BERT}_\mathrm{large}$~\cite{DBLP:conf/naacl/DevlinCLT19} & 46.99 & 53.74 & 46.89 & 53.32 & 49.27 & 56.54 & 61.63 \\
		$\mathrm{BERT}_\mathrm{large}$-last2avg~\cite{li2020sentence}  & 59.56 & 60.22 & 57.68 & 61.37 & 61.02 & 68.04 & 70.32 \\
		KSG k=10~\cite{zhelezniak2020estimating} & - & - & 60.40 & 61.50 & 68.30 & 77.00 & 75.10 \\
		MaxPool+KSG k=10~\cite{zhelezniak2020estimating} & - & - & 59.50 & 60.20 & 67.50 & 75.00 & 74.10\\
		DynaMax Jaccard~\cite{DBLP:conf/iclr/ZhelezniakSSMFH19}  & - & - & 61.30 & 61.70 & 66.90 & 76.50 & 74.70 \\
		CKA dCorr~\cite{DBLP:conf/emnlp/ZhelezniakSBSH19}  & - & - & 60.90 & 63.40 & 67.80 & 76.20 & 73.40 \\
		CKA Gaussian~\cite{DBLP:conf/emnlp/ZhelezniakSBSH19} & - & - & 60.80 & \textbf{64.60}  & 68.00 & 76.40 & 73.80 \\\hline
		ROTS+SWC+mean & \textbf{72.69} &  \textbf{62.88} &  \textbf{63.07} & 62.61 &  \textbf{70.73} & \textbf{78.06} & \textbf{75.74}\\
		ROTS+SUP+mean  &  71.63 &  61.81 &    62.13 &    61.04 &    70.85 &    77.26 &    74.50 \\
		ROTS+WR+mean & 69.78 &  61.39 &    61.48 &    59.29 &    70.19 &    75.18 &    73.26 \\\hline
	\end{tabular}
\end{table*}

We re-implement SIF, uSIF and WRD and compare the Pearson's $r\times 100$ in Table~\ref{tb:benchmark-related} together with the ACV-Tree\footnote{Scores extracted from~\cite{le2018acv}, STS13 is not valid since they didn't report on SMT subtask} and BERTScore+fastText\footnote{Scores extracted from~\cite{yokoi2020word}}.
Other baselines are compared by Spearman's $ \rho $ in Table~\ref{tb:benchmark-others}. 
The clear advantage of ROTS-mean is shown. 
Our results confirm the finding reported by~\cite{yokoi2020word} that the BERT-based method is sub-optimal under unsupervised settings.


\subsection{Ablation Study}\label{exp:ablation-study}

For ablation study, 
the scores are averaged from scores on the three datasets, including SICK, STSB test, and Twitter. 
We don't include STS12-16 since they overlap with STSB. \textit{Depths and Aggregations}, \textit{Correction and Prior}, and \textit{Recursive Phrase Partitions} are discussed since they are closely related to ROTS.
More experiments on different \textit{word vectors} and \textit{preprocessings} can be found in Appendix~\ref{app:extend-experiments}.
Uncertainty quantification by the BCa confidence interval~\cite{efron1987better} on different datasets can be found in Appendix~\ref{app:breakdown}.

\begin{figure}[t]
\centering
	\includegraphics[width=\linewidth]{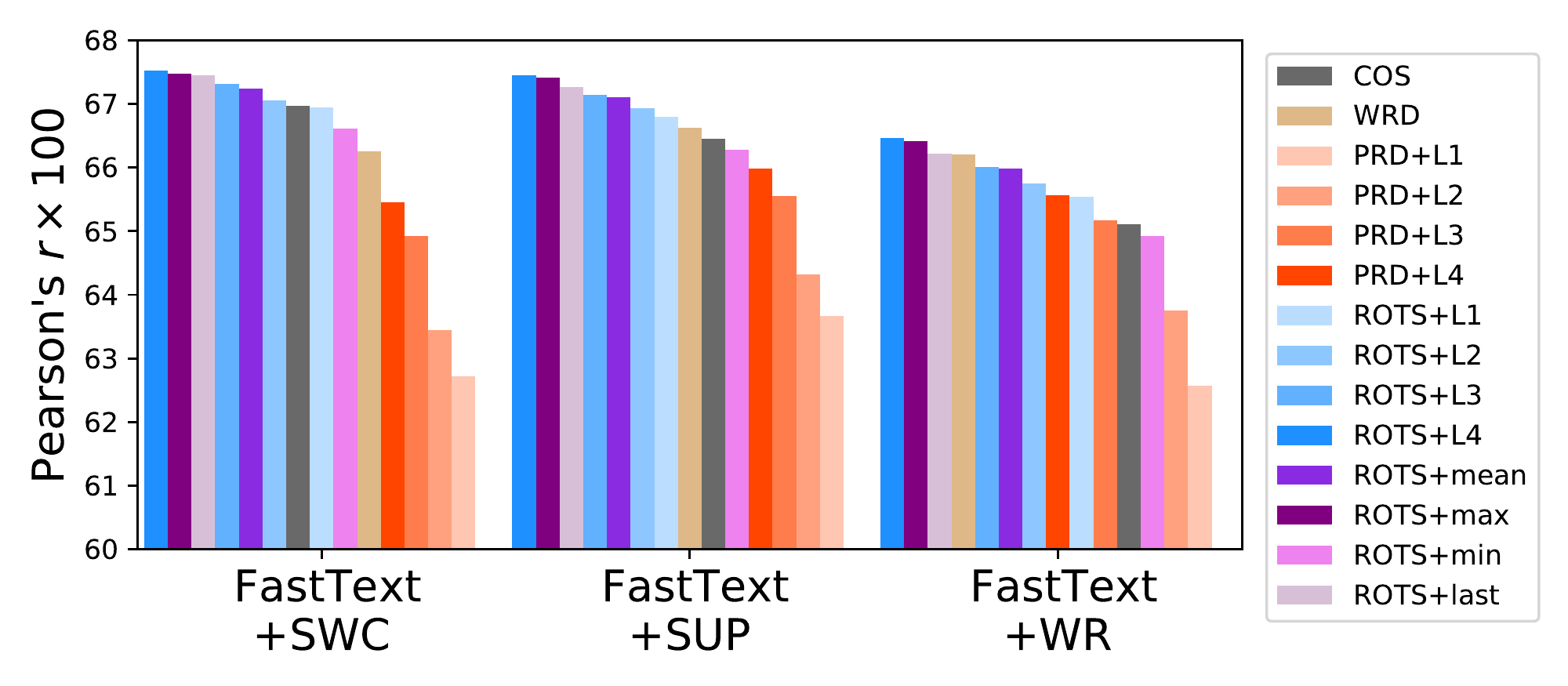}
	\caption{Ablation study for ROTS depth and aggregation. Scores are averaged from STSB, Twitter and SICK.}\label{fig:ablation-depth}
\end{figure}

\paragraph{Depths and Aggregations}
Once all $\mathrm{ROTS}_k$ are obtained, we consider different aggregation methods including \textit{mean}, \textit{max}, \textit{min}, \textit{last} and picks the $ k $-th level.
The ablation study of depth and aggregation is shown in Figure~\ref{fig:ablation-depth}.
We report the ROTS results at different levels and different aggregations. 
We also include the Phrase Rotator's Distance (PRD) at the same recursive phrase partitions as ROTS. 
PRD-L$ k $ is the special case of ROTS-L$ k $ by setting $ \epsilon_k=0 $ and $ \alpha=0 $. 
AC is equivalent to 0-th level ROTS and WRD is the L-th level of PRD so they are included.

ROTS similarities (blue and purple bars) dominate among all other baselines.
We can see that the performances of ROTS and PRD increase as their levels get deeper (the related bars are plotted with deeper blue and orange colors).
Interestingly, PRDs are generally worse than WRD, which indicates that the naive phrase alignment may not be suitable, and may suffer from sub-optimal inter-sentence alignment and intra-sentence semantics.
The performance gains of ROTS-L$ k $ from PRD-L$ k $ clearly show that both the coarse-to-fine prior and the EC similarity are important.

\begin{figure}[t]
	\includegraphics[width=\linewidth]{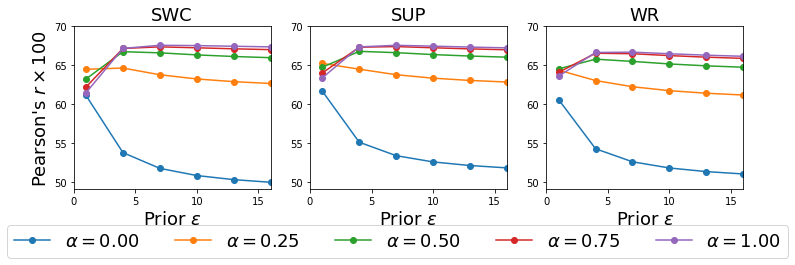}
	\caption{Effects of correction and prior for ROTS-L4 on fastText vectors}\label{fig:ec}
\end{figure}

\paragraph{Correction Step and Prior}
We adjust the $\alpha$ in Definition~\ref{def:ecformulation} to control the correction effect and the $\epsilon_k$ for the prior strength at the $ k $-th phrase partition level. 
For simplicity, we assume prior strengths $\epsilon_k$ are controlled by the single parameter $\epsilon$.
We report the Pearson's $r\times 100$ of ROTS-L4 averaged on the three datasets in Figure~\ref{fig:ec} since ROTS-L4 is the best in Figure~\ref{fig:ablation-depth}.
As shown in Figure~\ref{fig:ec}, proper correction and prior are essential to produce good performances. 
The correction step is very important since results without it decrease significantly. 
This is consistent with the PRD observation in Figure~\ref{fig:ablation-depth}. 
$\alpha$ and $\epsilon$ can be chosen easily since the performances is good and consistent if $\alpha > 0.5$ and $\epsilon > 5$.

\paragraph{Recursive Phrase Partitions}
ROTS relies on the recursive phrase partitions that might be produced from parsing trees. 
Instead of exhausting the parsers, we consider the simplest binary tree, i.e. the sub-phrase partition is constructed by uniformly splitting each phrase, to show the lower bound of the ROTS performances.
We see from Table~\ref{tb:parser} that the ROTS with spaCy dependency parser performs best in all cases among related baselines.
Given the preprocessing setups, we find that the binary tree still outperforms WRD and AC with SUP and SWC setups. 
For sub-optimal WR setup, ROTS with the binary tree are very close to that in WRD and better than AC.
Though preprocessing setups affect the performance, we can observe the performance gain by introducing the recursive phrase partitions given the setup.
Therefore, we conclude that the coarse-to-fine prior captures the intra-sentence structures. 
The performance gain can be observed by even the simplest binary tree.

\subsection{More empirical experiments}
Some results are presented in the Appendix, including the justification of more choices on preprocessing in Appendix~\ref{app:other-prep}, comparison under transfer and supervised setting in Appendix~\ref{app:extend}, computation time in Appendix~\ref{speed}, interpolation of WRD and AC by EC similarity in Appendix~\ref{interp}.

\begin{table}[t]
\centering
\caption{Pearson's $r\times 100$ for different parsers. spaCy: ROTS-L4 with the spaCy parser, Binary: ROTS-L4 with a binary tree. The best score is indicated in the boldface and the second highest score is underlined.}\label{tb:parser}
\small
\begin{tabular}{|l|ll|ll|}
    \hline
    Model & spaCy & Binary & WRD & AC\\\hline
    fastText + SUP & \textbf{67.45} & \underline{67.20} & 66.63 &66.45\\
    fastText + SWC & \textbf{67.52} & \underline{67.26} & 66.26 & 66.97\\
    fastText + WR  & \textbf{66.47} & 66.15 & \underline{66.20} & 65.11\\
    \hline
\end{tabular}\vskip-1em
\end{table}


\section{Conclusion}

In this paper, we present a new EC similarity of STS that allows flexible adaptation of word-level alignment, which successfully unifies three different unsupervised approaches.
By taking advantage of the recursive phrase partitions, we generalize EC similarity to the phrase alignment. 
Then, we propose ROTS, a new sentence similarity that considers phrase semantics by conducting phrase alignment in a coarse-to-fine order under the coarse-to-fine prior OT.
The thorough comparison with unsupervised baselines demonstrates the state-of-the-art performance and technical details of ROTS are fully justified by the ablation study.

\section{Acknowledgement}
This work was supported by the National Key R\&D Program of China (2020AAA0109603).

\bibliography{anthology,custom}
\bibliographystyle{acl_natbib}

\appendix
\clearpage

\section{Extended Experimental Setup information}

\subsection{Dataset details}~\label{app:dataset}
\begin{compactitem}
\item \textbf{STSB} dev and test set in STS-Benchmark~\cite{DBLP:journals/corr/abs-1708-00055}. It can be downloaded directly from \cite{DBLP:conf/rep4nlp/EthayarajhH18}'s implementation~\footnote{\url{https://github.com/kawine/usif}}.
\item \textbf{STS[year]} STS from 2012 to 2016~\cite{DBLP:conf/semeval/AgirreCDG12,DBLP:conf/starsem/AgirreCDGG13,DBLP:conf/semeval/AgirreBCCDGGMRW14,DBLP:conf/semeval/AgirreBCCDGGLMM15,DBLP:conf/semeval/AgirreBCDGMRW16}. 
STS12 contains 5 subtasks, STS13 contains 4 subtasks, STS14 contains 6 subtasks, STS15 contains 5 subtasks and STS16 contains 5 subtasks. 
The reported score is averaged from scores for related subtasks. 
It can be obtained by~\cite{conneau2018senteval}'s implementation~\footnote{\url{https://github.com/facebookresearch/SentEval}}. 
A newer implementation is also available~\footnote{\url{https://github.com/babylonhealth/corrsim}}.
\item \textbf{SICK} Semantic relatedness task at SemEval 2014~\cite{DBLP:conf/semeval/MarelliBBBMZ14}. 
It can be downloaded directly from \cite{DBLP:conf/rep4nlp/EthayarajhH18}'s implementation~\footnote{\url{https://github.com/kawine/usif}}.
\item \textbf{Twitter} Paraphrase and semantic similarity in Twitter (PIT) at SemEval 2015~\cite{DBLP:conf/semeval/XuCD15}. 
This dataset was obtained by emailing the author.
\end{compactitem}

\subsection{Pretrained Word Vectors}~\label{app:wv}
We list the downloadable links of word vectors used in this paper.
\begin{compactitem}
\item Word2Vec~\cite{DBLP:conf/nips/MikolovSCCD13}: We use the pretrained word2vec~\footnote{\url{GoogleNews-vectors-negative300.bin.gz}}. 
However, this file is in \texttt{.bin} format. 
We use gensim~\cite{rehurek_lrec} to convert the file to \texttt{.vec} format.
\item GloVe~\cite{DBLP:conf/emnlp/PenningtonSM14}: We use the 300D GloVe vectors trained on Common Crawl (840B tokens, 2.2M vocabulary)~\footnote{\url{http://nlp.stanford.edu/data/glove.840B.300d.zip}}.
\item fastText~\cite{bojanowski2017enriching}: We use the 300D fastText vectors trained on Common Crawl  (600B tokens) without subword information~\footnote{\url{https://dl.fbaipublicfiles.com/fasttext/vectors-english/crawl-300d-2M.vec.zip}}.
\item PSL~\cite{DBLP:journals/tacl/WietingBGL15}: We use the pretrained vectors from the author~\footnote{\url{https://drive.google.com/file/d/0B9w48e1rj-MOck1fRGxaZW1LU2M/view?usp=sharing}}.
\item ParaNMT~\cite{DBLP:conf/acl/GimpelW18}: Two versions are provided by the author~\footnote{\url{https://www.cs.cmu.edu/~jwieting/}} and we keep the same choice as ~\cite{DBLP:conf/rep4nlp/EthayarajhH18}~\footnote{\url{https://github.com/kawine/usif/blob/master/paranmt.tar.gz}}.
\end{compactitem}

\subsection{Preprocessing of word vectors}~\label{app:preprocessing}

Other preprocessing setups are discussed as follows:
Here we list several preprocessing approaches mentioned in previous research. For those with hyper-parameters, we also give the hyper-parameters used in this paper.

\begin{compactitem}
\item \textbf{Word-level} Each word is associated with one weight. We consider SIF (\textbf{W})eights~\cite{DBLP:conf/iclr/AroraLM17} with $a=10^{-3}$ and (\textbf{U})SIF weights~\cite{DBLP:conf/rep4nlp/EthayarajhH18}.
\item \textbf{Vocabulary-level} Vectors are modified based on the vectors of words in the whole vocabulary, e.g. (\textbf{A})ll-but-the-top~\cite{DBLP:conf/iclr/MuV18} with $D=3$, (\textbf{C})onceptor negation~\cite{DBLP:conf/naacl/LiuUS19} with $\alpha = 2$.
\item \textbf{Sentence-level} Vectors are modified by vectors of words in the same sentence, including Dimension-wise (\textbf{S})caling~\cite{DBLP:conf/rep4nlp/EthayarajhH18}.
\item \textbf{Corpus-level} Vectors are modified based on all sentences in the corpus, e.g. main component (\textbf{R})emoval~\cite{DBLP:conf/iclr/AroraLM17} and (\textbf{P})iece-wise component removal~\cite{DBLP:conf/rep4nlp/EthayarajhH18} with $p=5$.
\end{compactitem}

\begin{table*}[t]
\centering
\caption{Necessary resources of typical STS model}~\label{tb:resources}
\small
\begin{tabular}{|p{3cm}|llp{1cm}p{1cm}|p{1cm}p{1cm}p{1.2cm}p{1.2cm}|}
\hline
\multicolumn{1}{|c|}{\multirow{2}{*}{Model}} & \multicolumn{4}{c|}{Pretrain}    & \multicolumn{4}{c|}{Data}                         \\\cline{2-9}
\multicolumn{1}{|c|}{} & Parser & Weights & Word Vector & Language Model & Training texts & Training labels & Transferred texts & Transferred labels \\\hline
\textit{Unsupervised setting} & & & & & & & &\\
BERT layer embedding \cite{DBLP:conf/naacl/DevlinCLT19} & \xmark & \xmark & \xmark & \cmark & \xmark & \xmark & \xmark& \xmark
\\
BERTScore \cite{DBLP:conf/iclr/ZhangKWWA20} & \xmark & \xmark & \xmark & \cmark & \xmark & \xmark & \xmark& \xmark\\
Additive composition \cite{DBLP:conf/iclr/AroraLM17,DBLP:conf/rep4nlp/EthayarajhH18} & \xmark & \cmark & \cmark & \xmark & \xmark & \xmark & \xmark& \xmark\\ 
DynaMax-Jaccard \cite{DBLP:conf/iclr/ZhelezniakSSMFH19} & \xmark & \cmark & \cmark & \xmark & \xmark & \xmark & \xmark& \xmark\\
Center Kernel Alignment \cite{DBLP:conf/emnlp/ZhelezniakSBSH19}  & \xmark & \cmark & \cmark & \xmark & \xmark & \xmark & \xmark& \xmark\\
KSG cross entropy \cite{zhelezniak2020estimating}  & \xmark & \cmark & \cmark & \xmark & \xmark & \xmark & \xmark& \xmark\\
OT \cite{DBLP:conf/icml/KusnerSKW15,yokoi2020word} & \xmark & \cmark & \cmark & \xmark & \xmark & \xmark & \xmark& \xmark\\
ACV-Tree \cite{le2018acv} & \cmark & \cmark & \cmark & \xmark & \xmark & \xmark & \xmark& \xmark\\
ROTS (ours) & \cmark & \cmark & \cmark & \xmark & \xmark & \xmark & \xmark& \xmark\\
\hline
\textit{Transfer and domain adaptation settings} & & & & & & & & \\
SentenceBERT \cite{reimers2019sentence} & \xmark & \xmark & \xmark & \cmark & \xmark & \xmark & *NLI&  *NLI\\
BERT-Flow-*NLI \cite{li2020sentence} & \xmark & \xmark & \xmark & \cmark & \xmark & \xmark & *NLI& \xmark\\
BERT-Flow-*target \cite{li2020sentence}  & \xmark & \xmark & \xmark & \cmark & \cmark & \xmark & \xmark& \xmark \\
SimCSE-*NLI \cite{gao2021simcse}  & \xmark & \xmark & \xmark & \cmark & \xmark & \xmark & *NLI & *NLI \\\hline
\textit{Fine-tuning LM} & & & & & & & &\\
BERT-Finetune & \xmark & \xmark & \xmark & \cmark & \cmark & \cmark & \xmark& \xmark\\
\hline
\end{tabular}
\end{table*}

\subsection{Various STS and the required resources}~\label{app:baselines}
We summarize the usage of data and other resources of popular STS models in Table~\ref{tb:resources}.
The key difference between unsupervised settings and other settings is the usage of external data to further train the model.
We majorly consider the approaches that can be used without training.

\section{Extended Experiments}~\label{app:extend-experiments}

\subsection{Joint effect of word vectors and preprocessings} We investigate the effects of the word vectors in Table~\ref{table:wv-pre}. It has been shown that fastText is the best word vector for all three kinds of unsupervised STS regardless of the three pre-processing steps. Furthermore, ROTS performs best compared to AC and WRD when the fastText is chosen.

\begin{table}[t]
\centering
\caption{Pearson's $r \times 100$ for benchmark study with different word vectors and pre-processing setups. The best word vector achieved given the same pre-processing setup is indicated by boldface. The best pre-processing setup given the word vector is underlined.}~\label{table:wv-pre}
\small
\begin{tabular}{|l|l|l|l|l|}
\hline
\multirow{2}{*}{Pre-processing} & \multicolumn{1}{c|}{\multirow{2}{*}{Word Vectors}} & \multicolumn{3}{c|}{Similarity} \\\cline{3-5} & \multicolumn{1}{c|}{} & AC & WRD & ROTS \\\hline
\multirow{3}{*}{WR} & fastText & \textbf{65.11} &  \textbf{66.20} & \textbf{66.47} \\
& GloVe & 57.86 & 61.92 & 60.73\\
& Word2Vec & 57.35 & 59.52 & 58.68\\\hline
\multirow{3}{*}{SWC} & fastText & \underline{\textbf{66.97}} & \textbf{66.26} & \underline{\textbf{67.52}}\\
 & GloVe & \underline{66.57} & \underline{65.21} & \underline{66.95}\\
 & Word2Vec & \underline{60.21} & \underline{60.13} & \underline{60.66}\\\hline
\multirow{3}{*}{SUP} & fastText & \textbf{66.45} & \underline{\textbf{66.63}} & \textbf{67.45}\\
 & GloVe & 64.08 & 65.04 & 65.53\\
 & Word2Vec & 57.69 & 59.49 & 58.90\\\hline

\end{tabular}
\end{table}

\subsection{Other preprocessing setups}~\label{app:other-prep}
In Table~\ref{table:wv-pre}, it is also found that SWC is the best performed pre-processing setup in eight out of nine combinations of word vectors and unsupervised STSes. It is also shown that given the SWC as the preprocessing setup, ROTS performs best among three unsupervised STS for all three kinds of word vectors.
We also explore the combination of preprocessing setups from word level to corpus level. We provide preliminary results about the impact of preprocessing on ROTS in STSB test split with fastText vectors. We report the score of ROTS-L4 in Table~\ref{tb:setups}.

\begin{table}
\centering
\small
\caption{STSB test results by ROTS-L4 with fastText vectors with different preprocessing setups.}\label{tb:setups}
    \begin{tabular}{|l|l|l|}
    \hline
        Setup & Pearson's $r\times 100$ & BCa 95\% CI\\\hline
        +W & 72.59 & [70.02, 74.90]\\
        +WR$^*$ & 72.34 & [69.81, 74.66]\\
        +U & 72.62 & [70.01, 74.96]\\
        +SU & 74.52 & [71.90, 76.89]\\
        +SUP$^*$ & 74.69 & [72.01, 77.00]\\
        +SW & 74.78 & [72.31, 77.11]\\
        +SWP & 74.90 & [72.32, 77.08]\\
        +SWA & 74.97 & [72.47, 77.15]\\
        +SUA & 75.20 & [72.73, 77.43]\\
        +SWR & 75.21 & [72.75, 77.45]\\
        +SUR & 75.35 & [72.88, 77.53]\\
        +SWC$^*$ & 75.66 & [73.23, 77.84]\\
        +SUC & 75.73 & [73.25, 77.95]\\
        +SWRC & 75.80 & [73.33, 77.92]\\
        +SWRCA & 75.86 & [73.46, 78.11]\\
        +SURC & 75.89 & [73.39, 78.08]\\
        +SURCA & 75.94 & [73.49, 78.18]\\
        \hline
    \end{tabular}
\end{table}

By Table~\ref{tb:setups}, we suggest that \textbf{C} at the vocabulary level, \textbf{S} at the sentence level, \textbf{R} at the corpus level are beneficial. 
It is not clear which one of \textbf{U} or \textbf{W} in the word level for ROTS-L4 is more effective. 
As a result, we propose to combine the choices for vocabulary, sentence, and corpus levels, i.e. \textbf{SCR} for ROTS with \textbf{U} or \textbf{W}.
Moreover, we think the two preprocessing setups in the vocabulary level, i.e. \textbf{CA} can also be combined. 
The best performance of ROTS-L4 is achieved by \textbf{SURCA}.
The setups suggested by previous words are starred in the table, i.e. \textbf{SWC} for WRD~\cite{yokoi2020word}, \textbf{SUP}~\cite{DBLP:conf/rep4nlp/EthayarajhH18} and \textbf{WR}~\cite{DBLP:conf/iclr/AroraLM17} for AC.
Though they may not be the best choice for ROTS-L4, we argue the results presented are sufficient to reveal the advantage of ROTS over other related baselines under various setups.

\begin{table*}[t]
	\caption{Spearman's $\rho \times 100$ for different models in semisupervised and transferred setting}\label{tb:transfer}
	\centering
	\small
	\begin{tabular}{|l|l|l|l|l|l|l|l|}
		\hline
		Similarity               & STS-B & SICK & STS12 & STS13 & STS14 & STS15 & STS16 \\\hline
		\textit{ParaNMT Transfer} & & & & & & & \\
		ROTS+WR+mean (ParaNMT)           &     78.51 &  65.90 &    65.39 &    63.95 &    \textbf{75.41} &    79.90 &    77.86 \\
            ROTS+SWC+mean (ParaNMT)           &     78.65 &  65.29 &    64.88 &    62.08 &    74.24 &    79.16 &    76.38 \\\hline
		\textit{SNLI + MNLI transfer} & & & & & & & \\
		Sentence BERT(large)     & \textbf{79.23}     & 73.75  & 72.27    & 78.46    & 74.90    & 80.99    & 76.25    \\
		Sentence RoBERTa(large)  & 79.10     & \textbf{74.29}  & \textbf{74.53}    & 77.00    & 73.18    & 81.85    & 76.82    \\\hline
		\textit{domain adaptation setting} & & & & & & & \\
		BERT (large) Flow *NLI   & 68.09     & 64.62  & 61.72    & 66.05    & 66.34    & 74.87    & 74.47    \\
		BERT (large) Flow target & 72.26     & 62.50  & 65.20    & 73.39    & 69.42    & 74.92    & 77.63 \\
		SimCSE-BERT *NLI & 76.85 & 72.23 & 68.40 & \textbf{82.41} & 74.38 & \textbf{80.91} & \textbf{78.56} \\
		\hline
	\end{tabular}
\end{table*}

\subsection{Evaluation for transfer and semi-supervised settings}\label{app:extend}

The results can be found in Table~\ref{tb:transfer}.
ROTS that using the transferred ParaNMT word vector has good performance even compared to Sentence BERT with pretrained BERT large or RoBERTa large~\cite{reimers2019sentence}, and is better than the domain adaptation settings~\cite{li2020sentence}.
It is shown that PLM based models~\cite{DBLP:conf/emnlp/CerYKHLJCGYTSK18,reimers2019sentence,li2020sentence,gao2021simcse} are on par with ROTS with transfered word vectors~\cite{DBLP:conf/acl/GimpelW18}.

\section{Computation Speed}\label{speed}

We report the computation speed for different similarities on a computer with an Intel i7 CPU of 2.6 GHz with 6 cores and 16 GB RAM.
The optimal transport is computed by the POT~\cite{flamary2017pot} package~\footnote{\url{https://github.com/PythonOT/POT}}.

We compare the computation of ROTS with WRD and PRD on STS-B test split (1379 sentence pairs to compute in total).
Notably, we focus on the speed by Sinkhorn algorithm~\cite{DBLP:conf/nips/Cuturi13} for two reasons: (1) it has $O(n^2)$ time complexity; (2) it can be easily accelerated by GPU.

Table~\ref{tb:speed} reports the speed by different OT-based algorithms. 
We note that the reported speeds for phrase alignment algorithms (PRD and ROTS) also include the time for parsing and constructing the recursive phrase partitions. 
This additional process brings additional computational overhead and slows down the speed. 
As a consequence of parsing, we can see that for PRD, \#OT/sec is slowed down compared to WRD. 
However, ROTS is based on Prior OT with larger regularization strength, and each call of \texttt{ot.sinkhorn} requires much less time, thus making up the computational overhead by parsing.

\begin{table*}[t]
\centering
\small
\caption{Comparison of computation speed}\label{tb:speed}
    \begin{tabular}{|l|lll|lll|}
    \hline
        Method & Function in POT & Reg & Reg. Strength & \#OT/STS & \#STS/sec & \#OT/sec \\\hline
        WRD & \texttt{ot.sinkhorn} & Entropy & 0.1 & 1 & 208.52 & 208.52\\
        PRD 4 levels & \texttt{ot.sinkhorn} & Entropy & 0.1 & 5 & 32.80 & 164.00\\
        ROTS 4 levels & \texttt{ot.sinkhorn} & KL Prior & 10 & 5 & 60.56 & 302.80\\
    \hline
    \end{tabular}
\end{table*}

\section{EC Interpolation of WRD and AC}\label{interp}


\begin{figure*}[t]
	\centering
	\includegraphics[trim=0.4cm 0.4cm 2cm 0.4cm, clip, width=\linewidth]{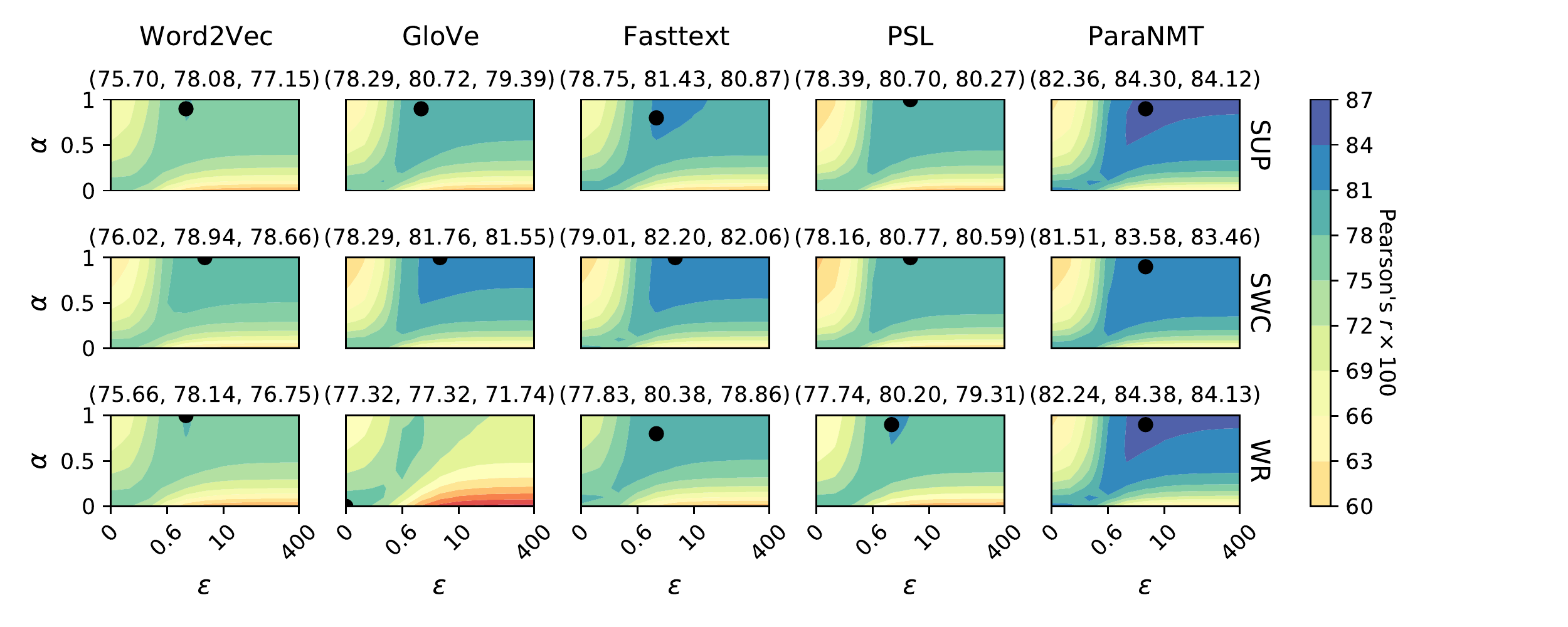}
	\caption{Interpolating AC and WRD by Example~\ref{eg:ec-interp} on STS-B dev set. For each case, black dots indicate the highest score. The title of each subplot indicates scores by (WRD, Best Interpolation, AC).}\label{fig:interp}
\end{figure*}

We consider 15 combinations from 5-word vectors and 3 preprocessing setups.
For each case, we grid-search 10 values of $\alpha$ by linearly splitting $[0,1]$, and 10 values of $\epsilon$ by logarithmically splitting $[0,400]$, resulting in 100 runs.
The performances of 15 cases are shown in Figure~\ref{fig:interp}. 
Grid-search results indicate that the proper choice of EC similarity outperforms both WRD and AC, thus showing solidness.
Specifically, most of the best interpolation performances appear when $\alpha = 1$ (14 cases) and $\epsilon\in [1, 10]$ (14 cases), which confirms the ablation study in Figure~\ref{fig:ec}.
This observation demonstrates the effectiveness of the correction term and indicates that the best choice of the alignment matrix should be chosen carefully.

\section{Dataset Breakdown Tables and Uncentity Quantification}\label{app:breakdown}

We provide the breakdown tables related to Figure~\ref{fig:ablation-depth} with different word vectors, e.g. fastText in Table~\ref{app:tb:fastText}, GloVe in Table~\ref{app:tb:GloVe}, Word2Vec in Table~\ref{app:tb:Word2Vec}, PSL in Table~\ref{app:tb:PSL} and ParaNMT in Table~\ref{app:tb:ParaNMT}.
We see that WRD performs consistently well on Twitter dataset. 
For STSB and SICK, ROTS is better, resulting in the best overall performance.

\subsection{Three Typical Preprocessings}\label{app:benchmark-uq}
We provide further information for Table~\ref{tb:benchmark-related}, including the Pearson's $r\times 100$ for each individual datasets, plus STSB dev split and Twitter.
Still, we focus on fastText vectors, and list three preprocessing setups, e.g. WR in Table~\ref{app:table:bm:WR}, SWC in Table~\ref{app:table:bm:SWC}, SUP in Table~\ref{app:table:bm:SUP}.
We find that ROTS has the best performance in WR and SUP, which is consistent with Table~\ref{tb:benchmark-related}, and  AC is good with SWC.

\begin{table*}[t]
\centering
\caption{Breakdown table for benchmark study with WR preprocessing}~\label{app:table:bm:WR}
\scriptsize
\begin{tabular}{|l|l|p{1cm}l|p{1cm}l|p{1cm}l|}
\hline
\multirow{2}{*}{dataset} & \multicolumn{1}{c|}{\multirow{2}{*}{subsplit/subtask}} & \multicolumn{2}{c|}{AC} & \multicolumn{2}{c|}{WRD} & \multicolumn{2}{c|}{ROTS} \\
 & \multicolumn{1}{c|}{} & Pearson's $r\times 100$ & BCa 95\% CI & Pearson's $r\times 100$ & BCa 95\% CI & Pearson's $r\times 100$ & BCa 95\% CI \\\hline
\multirow{2}{*}{STSB} & test & 70.13 & {[}67.35, 72.53{]} & 73.13 & {[}70.40, 75.54{]} & 72.34 & {[}69.78, 74.66{]} \\
 & dev & 78.85 & {[}76.93, 80.56{]} & 77.83 & {[}75.49, 79.87{]} & 79.78 & {[}77.76, 81.48{]} \\\hline
Twitter & test & 52.01 & {[}47.67, 56.15{]} & 56.73 & {[}52.07, 60.83{]} & 53.32 & {[}48.75, 57.45{]} \\\hline
SICK & test & 73.20 & {[}71.66, 74.68{]} & 68.73 & {[}67.13, 70.27{]} & 73.75 & {[}72.16, 75.17{]} \\\hline
\multirow{5}{*}{STS12} & MSRpar & 39.84 & {[}33.68, 45.44{]} & 49.93 & {[}44.04, 54.97{]} & 43.04 & {[}36.93, 48.40{]} \\
 & MSRvid & 86.01 & {[}84.22, 87.59{]} & 82.38 & {[}79.82, 84.57{]} & 86.35 & {[}84.59, 87.91{]} \\
 & SMTeuroparl & 52.41 & {[}44.52, 60.64{]} & 52.01 & {[}45.61, 58.37{]} & 51.95 & {[}44.82, 59.52{]} \\
 & OnWN & 74.31 & {[}70.55, 77.55{]} & 74.59 & {[}71.42, 77.30{]} & 74.43 & {[}70.78, 77.58{]} \\
 & SMTnews & 64.71 & {[}54.76, 73.79{]} & 60.13 & {[}52.37, 67.16{]} & 62.92 & {[}53.37, 72.18{]} \\\hline
\multirow{4}{*}{STS13} & FNWN & 42.94 & {[}29.66, 53.80{]} & 48.59 & {[}35.52, 58.61{]} & 44.64 & {[}31.36, 55.42{]} \\
 & headlines & 72.88 & {[}69.45, 75.99{]} & 72.49 & {[}68.53, 75.73{]} & 73.52 & {[}69.99, 76.54{]} \\
 & OnWN & 82.36 & {[}79.81, 84.57{]} & 69.74 & {[}65.22, 73.64{]} & 80.67 & {[}77.85, 83.07{]} \\
 & SMT & 39.03 & {[}31.69, 46.65{]} & 41.56 & {[}34.95, 47.44{]} & 40.65 & {[}33.25, 47.89{]} \\\hline
\multirow{6}{*}{STS14} & deft-forum & 51.99 & {[}45.09, 58.05{]} & 46.53 & {[}38.71, 53.75{]} & 50.78 & {[}43.96, 56.97{]} \\
 & deft-news & 74.54 & {[}68.55, 78.93{]} & 74.62 & {[}68.82, 79.55{]} & 75.52 & {[}69.80, 79.83{]} \\
 & headlines & 68.71 & {[}64.55, 72.33{]} & 67.29 & {[}62.68, 71.45{]} & 69.27 & {[}65.17, 73.02{]} \\
 & OnWN & 84.52 & {[}82.31, 86.32{]} & 76.45 & {[}73.25, 79.17{]} & 83.62 & {[}81.32, 85.56{]} \\
 & images & 81.33 & {[}78.78, 83.43{]} & 80.06 & {[}77.09, 82.50{]} & 81.72 & {[}79.20, 83.82{]} \\
 & tweet-news & 76.62 & {[}72.87, 79.78{]} & 78.65 & {[}75.59, 81.33{]} & 78.27 & {[}74.88, 81.19{]} \\\hline
\multirow{5}{*}{STS15} & answers-forums & 70.51 & {[}64.96, 75.13{]} & 75.15 & {[}70.11, 79.29{]} & 71.86 & {[}66.44, 76.29{]} \\
 & answers-students & 70.86 & {[}66.85, 74.32{]} & 76.02 & {[}72.59, 79.00{]} & 72.49 & {[}68.70, 75.71{]} \\
 & belief & 68.88 & {[}61.09, 74.18{]} & 77.71 & {[}71.39, 81.99{]} & 70.61 & {[}62.59, 75.62{]} \\
 & headlines & 74.44 & {[}71.27, 77.11{]} & 73.69 & {[}70.17, 76.70{]} & 74.79 & {[}71.52, 77.53{]} \\
 & images & 81.68 & {[}79.23, 83.74{]} & 83.83 & {[}81.28, 85.90{]} & 83.17 & {[}80.75, 85.13{]} \\\hline
\multirow{5}{*}{STS16} & answer-answer & 47.15 & {[}37.71, 55.50{]} & 60.61 & {[}52.21, 67.26{]} & 53.36 & {[}44.49, 60.90{]} \\
 & headlines & 72.39 & {[}66.12, 77.41{]} & 73.41 & {[}65.76, 79.02{]} & 73.55 & {[}67.06, 78.37{]} \\
 & plagiarism & 82.01 & {[}77.92, 85.28{]} & 82.46 & {[}77.73, 86.24{]} & 82.53 & {[}78.53, 85.90{]} \\
 & postediting & 79.37 & {[}71.53, 83.43{]} & 86.11 & {[}81.33, 89.16{]} & 79.81 & {[}72.99, 83.71{]} \\
 & question-question & 73.03 & {[}66.30, 77.87{]} & 69.79 & {[}61.18, 76.30{]} & 73.79 & {[}67.24, 78.60{]} \\\hline
\multicolumn{2}{|c|}{MEAN} & 68.51 & - & 69.32 & - & \textbf{69.40} & -\\
\hline
\end{tabular}
\end{table*}

\begin{table*}[t]
\centering
\caption{Breakdown table for benchmark study with SWC preprocessing}~\label{app:table:bm:SWC}
\scriptsize
\begin{tabular}{|l|l|p{1cm}l|p{1cm}l|p{1cm}l|}
\hline
\multirow{2}{*}{dataset} & \multicolumn{1}{c|}{\multirow{2}{*}{subsplit/subtask}} & \multicolumn{2}{c|}{AC} & \multicolumn{2}{c|}{WRD} & \multicolumn{2}{c|}{ROTS} \\
 & \multicolumn{1}{c|}{} & Pearson's $r\times 100$ & BCa 95\% CI & Pearson's $r\times 100$ & BCa 95\% CI & Pearson's $r\times 100$ & BCa 95\% CI \\\hline
\multirow{2}{*}{STSB} & test & 74.78 & {[}72.31, 77.06{]} & 74.58 & {[}72.08, 76.81{]} & 75.66 & {[}73.17, 77.86{]} \\
 & dev & 82.06 & {[}80.25, 83.67{]} & 78.47 & {[}76.27, 80.45{]} & 81.33 & {[}79.36, 83.08{]} \\\hline
Twitter & test & 54.01 & {[}49.35, 58.23{]} & 57.10 & {[}52.42, 61.23{]} & 55.56 & {[}50.95, 59.88{]} \\\hline
SICK & test & 72.12 & {[}70.63, 73.57{]} & 67.09 & {[}65.49, 68.60{]} & 71.33 & {[}69.80, 72.76{]} \\\hline
\multirow{5}{*}{STS12} & MSRpar & 52.05 & {[}46.24, 57.40{]} & 54.85 & {[}49.29, 59.92{]} & 51.96 & {[}46.18, 57.32{]} \\
 & MSRvid & 87.23 & {[}85.39, 88.76{]} & 80.91 & {[}78.23, 83.20{]} & 85.48 & {[}83.36, 87.22{]} \\
 & SMTeuroparl & 55.44 & {[}49.45, 61.30{]} & 52.75 & {[}46.66, 58.26{]} & 52.94 & {[}46.55, 59.00{]} \\
 & OnWN & 73.66 & {[}69.82, 77.04{]} & 73.80 & {[}70.63, 76.62{]} & 73.52 & {[}69.84, 76.80{]} \\
 & SMTnews & 56.28 & {[}47.71, 64.54{]} & 56.68 & {[}49.28, 63.64{]} & 54.27 & {[}46.13, 62.82{]} \\\hline
\multirow{4}{*}{STS13} & FNWN & 53.69 & {[}41.93, 62.83{]} & 47.98 & {[}36.27, 57.19{]} & 53.49 & {[}41.82, 63.02{]} \\
 & headlines & 75.92 & {[}72.66, 78.76{]} & 73.67 & {[}70.14, 76.79{]} & 75.38 & {[}71.92, 78.32{]} \\
 & OnWN & 82.89 & {[}80.04, 85.20{]} & 67.57 & {[}62.91, 71.77{]} & 76.66 & {[}73.17, 79.70{]} \\
 & SMT & 41.81 & {[}34.69, 48.45{]} & 40.98 & {[}34.52, 46.79{]} & 42.44 & {[}35.82, 48.50{]} \\\hline
\multirow{6}{*}{STS14} & deft-forum & 55.57 & {[}48.68, 61.79{]} & 48.98 & {[}41.00, 55.68{]} & 54.31 & {[}47.11, 60.52{]} \\
 & deft-news & 75.92 & {[}70.33, 80.46{]} & 75.63 & {[}70.18, 80.18{]} & 76.13 & {[}70.56, 80.70{]} \\
 & headlines & 71.27 & {[}67.27, 74.83{]} & 69.07 & {[}64.70, 73.00{]} & 71.13 & {[}67.08, 74.74{]} \\
 & OnWN & 85.05 & {[}82.85, 86.93{]} & 75.28 & {[}72.13, 78.04{]} & 81.41 & {[}78.77, 83.66{]} \\
 & images & 83.08 & {[}80.47, 85.19{]} & 79.24 & {[}76.29, 81.77{]} & 82.02 & {[}79.39, 84.25{]} \\
 & tweet-news & 79.02 & {[}75.83, 81.82{]} & 78.16 & {[}75.11, 80.76{]} & 79.30 & {[}76.24, 82.05{]} \\\hline
\multirow{5}{*}{STS15} & answers-forums & 75.46 & {[}70.52, 79.33{]} & 75.29 & {[}70.37, 79.42{]} & 75.76 & {[}70.83, 79.78{]} \\
 & answers-students & 74.15 & {[}70.73, 77.18{]} & 76.29 & {[}73.12, 79.10{]} & 74.10 & {[}70.68, 77.12{]} \\
 & belief & 78.22 & {[}72.24, 82.23{]} & 77.92 & {[}72.15, 82.19{]} & 78.56 & {[}73.04, 82.59{]} \\
 & headlines & 77.10 & {[}74.12, 79.68{]} & 75.11 & {[}71.84, 78.09{]} & 76.73 & {[}73.70, 79.52{]} \\
 & images & 85.48 & {[}83.09, 87.40{]} & 83.65 & {[}80.95, 85.76{]} & 85.39 & {[}83.04, 87.25{]} \\\hline
\multirow{5}{*}{STS16} & answer-answer & 60.44 & {[}51.81, 67.64{]} & 63.92 & {[}56.17, 70.26{]} & 61.66 & {[}53.52, 68.76{]} \\
 & headlines & 75.61 & {[}68.93, 80.37{]} & 75.28 & {[}67.87, 80.65{]} & 75.71 & {[}68.87, 80.67{]} \\
 & plagiarism & 83.48 & {[}79.25, 86.80{]} & 81.55 & {[}76.49, 85.31{]} & 82.46 & {[}77.98, 86.01{]} \\
 & postediting & 83.63 & {[}77.69, 87.06{]} & 86.90 & {[}82.05, 89.85{]} & 84.28 & {[}78.84, 87.61{]} \\
 & question-question & 76.09 & {[}68.65, 81.22{]} & 69.66 & {[}61.28, 76.39{]} & 76.08 & {[}68.90, 81.01{]} \\\hline
\multicolumn{2}{|c|}{MEAN} & 71.78 & - & 69.60 & - & \textbf{71.21} & -\\
\hline
\end{tabular}
\end{table*}

\begin{table*}[t]
\centering
\caption{Breakdown table for benchmark study with SUP preprocessing}~\label{app:table:bm:SUP}
\scriptsize
\begin{tabular}{|l|l|p{1cm}l|p{1cm}l|p{1cm}l|}
\hline
\multirow{2}{*}{dataset} & \multicolumn{1}{c|}{\multirow{2}{*}{subsplit/subtask}} & \multicolumn{2}{c|}{AC} & \multicolumn{2}{c|}{WRD} & \multicolumn{2}{c|}{ROTS} \\
 & \multicolumn{1}{c|}{} & Pearson's $r\times 100$ & BCa 95\% CI & Pearson's $r\times 100$ & BCa 95\% CI & Pearson's $r\times 100$ & BCa 95\% CI \\\hline
\multirow{2}{*}{STSB} & test & 73.47 & {[}70.90, 75.86{]} & 74.80 & {[}72.27, 77.13{]} & 74.69 & {[}72.12, 77.01{]} \\
 & dev & 80.87 & {[}78.98, 82.55{]} & 78.75 & {[}76.54, 80.72{]} & 81.15 & {[}79.23, 82.92{]} \\\hline
Twitter & test & 53.15 & {[}48.54, 57.23{]} & 57.41 & {[}52.87, 61.57{]} & 54.88 & {[}50.28, 59.08{]} \\\hline
SICK & test & 72.73 & {[}71.16, 74.24{]} & 67.67 & {[}66.06, 69.17{]} & 72.77 & {[}71.26, 74.22{]} \\\hline
\multirow{5}{*}{STS12} & MSRpar & 41.40 & {[}35.15, 46.68{]} & 50.98 & {[}45.25, 56.10{]} & 44.77 & {[}38.66, 50.10{]} \\
 & MSRvid & 86.79 & {[}84.94, 88.35{]} & 83.27 & {[}80.83, 85.32{]} & 87.04 & {[}85.18, 88.62{]} \\
 & SMTeuroparl & 53.29 & {[}46.42, 60.68{]} & 52.73 & {[}46.49, 58.45{]} & 52.31 & {[}45.42, 58.92{]} \\
 & OnWN & 73.53 & {[}69.66, 76.91{]} & 73.85 & {[}70.76, 76.55{]} & 73.56 & {[}69.81, 76.79{]} \\
 & SMTnews & 61.19 & {[}51.71, 70.15{]} & 59.34 & {[}52.19, 66.10{]} & 59.92 & {[}50.88, 69.08{]} \\\hline
\multirow{4}{*}{STS13} & FNWN & 49.52 & {[}37.76, 59.05{]} & 49.16 & {[}37.65, 58.60{]} & 50.25 & {[}38.73, 59.72{]} \\
 & headlines & 73.73 & {[}70.24, 76.79{]} & 72.95 & {[}69.30, 76.21{]} & 74.11 & {[}70.65, 77.17{]} \\
 & OnWN & 83.15 & {[}80.48, 85.50{]} & 71.12 & {[}66.69, 74.82{]} & 81.14 & {[}78.13, 83.53{]} \\
 & SMT & 39.22 & {[}32.18, 46.23{]} & 40.78 & {[}34.34, 46.68{]} & 40.50 & {[}33.48, 47.03{]} \\\hline
\multirow{6}{*}{STS14} & deft-forum & 53.39 & {[}46.27, 59.71{]} & 47.60 & {[}39.96, 54.72{]} & 52.21 & {[}45.10, 58.70{]} \\
 & deft-news & 76.08 & {[}70.28, 80.42{]} & 75.38 & {[}69.79, 80.11{]} & 76.91 & {[}71.41, 81.28{]} \\
 & headlines & 69.86 & {[}65.85, 73.29{]} & 68.11 & {[}63.55, 72.19{]} & 70.12 & {[}65.97, 73.73{]} \\
 & OnWN & 85.37 & {[}83.20, 87.16{]} & 77.55 & {[}74.59, 80.08{]} & 84.05 & {[}81.72, 85.98{]} \\
 & images & 83.73 & {[}81.36, 85.71{]} & 81.05 & {[}78.26, 83.51{]} & 83.65 & {[}81.31, 85.67{]} \\
 & tweet-news & 77.78 & {[}74.24, 80.79{]} & 78.24 & {[}75.16, 80.85{]} & 78.86 & {[}75.59, 81.61{]} \\\hline
\multirow{5}{*}{STS15} & answers-forums & 74.60 & {[}69.78, 78.61{]} & 75.80 & {[}70.74, 79.72{]} & 75.40 & {[}70.34, 79.27{]} \\
 & answers-students & 70.66 & {[}66.65, 74.13{]} & 75.16 & {[}71.79, 78.21{]} & 72.18 & {[}68.21, 75.47{]} \\
 & belief & 76.53 & {[}70.16, 80.98{]} & 78.37 & {[}72.23, 82.48{]} & 77.23 & {[}70.94, 81.64{]} \\
 & headlines & 75.16 & {[}72.06, 77.80{]} & 74.28 & {[}70.94, 77.36{]} & 75.42 & {[}72.23, 78.15{]} \\
 & images & 84.70 & {[}82.47, 86.60{]} & 84.66 & {[}82.23, 86.64{]} & 85.31 & {[}83.12, 87.13{]} \\\hline
\multirow{5}{*}{STS16} & answer-answer & 54.75 & {[}45.45, 62.55{]} & 62.77 & {[}54.87, 69.23{]} & 58.21 & {[}49.27, 65.50{]} \\
 & headlines & 72.71 & {[}66.44, 77.66{]} & 74.16 & {[}66.53, 79.74{]} & 73.90 & {[}67.05, 78.90{]} \\
 & plagiarism & 82.37 & {[}78.29, 85.55{]} & 82.21 & {[}77.61, 86.05{]} & 82.62 & {[}78.53, 85.92{]} \\
 & postediting & 82.64 & {[}76.07, 86.31{]} & 86.54 & {[}81.65, 89.56{]} & 83.08 & {[}76.90, 86.66{]} \\
 & question-question & 74.91 & {[}67.20, 79.98{]} & 71.21 & {[}63.00, 77.61{]} & 75.57 & {[}68.48, 80.59{]} \\
\hline\multicolumn{2}{|c|}{MEAN} & 70.25 & - & 69.86 & - & \textbf{70.75} & -\\
\hline
\end{tabular}
\end{table*}

\subsection{Five Word Vectors}

\begin{table*}[]
\centering
\caption{Breakdown table with fastText vectors}~\label{app:tb:fastText}
\scriptsize
\begin{tabular}{|l|p{1cm}l|p{1cm}l|p{1cm}l|p{1cm}|}
\hline
Similarity            & \multicolumn{2}{c|}{STSB}                           & \multicolumn{2}{c|}{Twitter}                        & \multicolumn{2}{c|}{SICK}                           & MEAN                          \\
             & Pearson's $r\times 100$                     & BCa 95\% CI        & Pearson's $r\times 100$                     & BCa 95\% CI        & Pearson's $r\times 100$                     & BCa 95\% CI        & Pearson's $r\times 100$                  \\\hline
with SUP &  &  &  &  &  &  &  \\
WRD & \cellcolor[HTML]{FCAF79}74.80 & {[}72.27, 77.13{]} & \cellcolor[HTML]{F8696B}57.41 & {[}52.87, 61.57{]} & \cellcolor[HTML]{73C27B}67.67 & {[}66.06, 69.17{]} & \cellcolor[HTML]{FFEB84}66.63 \\
AC & \cellcolor[HTML]{E7E482}73.47 & {[}70.90, 75.86{]} & \cellcolor[HTML]{D6DF81}53.15 & {[}48.54, 57.23{]} & \cellcolor[HTML]{FEEA83}72.73 & {[}71.16, 74.24{]} & \cellcolor[HTML]{EEE683}66.45 \\
ROTS+L0 & \cellcolor[HTML]{E7E482}73.47 & {[}70.77, 75.76{]} & \cellcolor[HTML]{D6DF81}53.15 & {[}48.47, 57.13{]} & \cellcolor[HTML]{FEEA83}72.73 & {[}71.19, 74.25{]} & \cellcolor[HTML]{EEE683}66.45 \\
ROTS+L1 & \cellcolor[HTML]{F3E783}73.77 & {[}71.24, 76.08{]} & \cellcolor[HTML]{E3E382}53.34 & {[}48.84, 57.51{]} & \cellcolor[HTML]{FCB179}73.31 & {[}71.71, 74.74{]} & \cellcolor[HTML]{FED17F}66.81 \\
ROTS+L2 & \cellcolor[HTML]{FDEA83}74.02 & {[}71.40, 76.34{]} & \cellcolor[HTML]{E7E482}53.40 & {[}48.76, 57.63{]} & \cellcolor[HTML]{FCAB78}73.37 & {[}71.85, 74.81{]} & \cellcolor[HTML]{FDBF7C}66.93 \\
ROTS+L3 & \cellcolor[HTML]{FFEB84}74.05 & {[}71.25, 76.44{]} & \cellcolor[HTML]{FFDA81}54.23 & {[}49.76, 58.44{]} & \cellcolor[HTML]{FDC47D}73.13 & {[}71.56, 74.58{]} & \cellcolor[HTML]{FBA176}67.14 \\
ROTS+L4 & \cellcolor[HTML]{FDB87B}74.69 & {[}72.12, 77.01{]} & \cellcolor[HTML]{FDC37D}54.88 & {[}50.28, 59.08{]} & \cellcolor[HTML]{FFE884}72.77 & {[}71.26, 74.22{]} & \cellcolor[HTML]{F9746D}67.45 \\
ROTS+mean & \cellcolor[HTML]{FFDB81}74.25 & {[}71.70, 76.56{]} & \cellcolor[HTML]{FFE583}53.93 & {[}49.50, 58.13{]} & \cellcolor[HTML]{FDC47D}73.13 & {[}71.59, 74.58{]} & \cellcolor[HTML]{FCA677}67.10 \\
ROTS+max & \cellcolor[HTML]{FDC17C}74.58 & {[}71.87, 76.86{]} & \cellcolor[HTML]{FDC37D}54.88 & {[}50.39, 59.04{]} & \cellcolor[HTML]{FFE884}72.77 & {[}71.27, 74.23{]} & \cellcolor[HTML]{F9796E}67.41 \\
ROTS+min & \cellcolor[HTML]{E3E382}73.36 & {[}70.57, 75.71{]} & \cellcolor[HTML]{BBD780}52.74 & {[}48.17, 56.90{]} & \cellcolor[HTML]{FFEB84}72.74 & {[}71.15, 74.17{]} & \cellcolor[HTML]{DFE182}66.28 \\
ROTS+last & \cellcolor[HTML]{FED781}74.30 & {[}71.70, 76.64{]} & \cellcolor[HTML]{FDC17C}54.95 & {[}50.51, 59.01{]} & \cellcolor[HTML]{F9E983}72.54 & {[}70.96, 74.03{]} & \cellcolor[HTML]{FA8F73}67.26 \\\hline
with SWC &  &  &  &  &  &  &  \\
WRD & \cellcolor[HTML]{FDC17C}74.58 & {[}72.08, 76.81{]} & \cellcolor[HTML]{F9746E}57.10 & {[}52.42, 61.23{]} & \cellcolor[HTML]{63BE7B}67.09 & {[}65.49, 68.60{]} & \cellcolor[HTML]{DDE182}66.26 \\
AC & \cellcolor[HTML]{FCB179}74.78 & {[}72.31, 77.06{]} & \cellcolor[HTML]{FFE283}54.01 & {[}49.35, 58.23{]} & \cellcolor[HTML]{EDE683}72.12 & {[}70.63, 73.57{]} & \cellcolor[HTML]{FDB97B}66.97 \\
ROTS+L0 & \cellcolor[HTML]{FDBF7C}74.60 & {[}72.08, 76.92{]} & \cellcolor[HTML]{FFEB84}53.74 & {[}49.07, 57.98{]} & \cellcolor[HTML]{DEE182}71.57 & {[}70.01, 73.01{]} & \cellcolor[HTML]{FFEA84}66.64 \\
ROTS+L1 & \cellcolor[HTML]{FCAC78}74.84 & {[}72.38, 77.09{]} & \cellcolor[HTML]{FFE383}53.99 & {[}49.38, 58.26{]} & \cellcolor[HTML]{EBE582}72.02 & {[}70.46, 73.44{]} & \cellcolor[HTML]{FDBC7B}66.95 \\
ROTS+L2 & \cellcolor[HTML]{FB9875}75.08 & {[}72.64, 77.42{]} & \cellcolor[HTML]{FFDF82}54.09 & {[}49.38, 58.41{]} & \cellcolor[HTML]{EAE582}72.00 & {[}70.50, 73.40{]} & \cellcolor[HTML]{FCAD78}67.06 \\
ROTS+L3 & \cellcolor[HTML]{FA8B72}75.25 & {[}72.69, 77.57{]} & \cellcolor[HTML]{FDC17C}54.95 & {[}50.26, 59.19{]} & \cellcolor[HTML]{E2E282}71.72 & {[}70.23, 73.17{]} & \cellcolor[HTML]{FA8871}67.31 \\
ROTS+L4 & \cellcolor[HTML]{F8696B}75.66 & {[}73.17, 77.86{]} & \cellcolor[HTML]{FCAB78}55.56 & {[}50.95, 59.88{]} & \cellcolor[HTML]{D8DF81}71.33 & {[}69.80, 72.76{]} & \cellcolor[HTML]{F8696B}67.52 \\
ROTS+mean & \cellcolor[HTML]{FA8471}75.33 & {[}72.82, 77.56{]} & \cellcolor[HTML]{FECD7F}54.59 & {[}49.93, 59.00{]} & \cellcolor[HTML]{E4E382}71.79 & {[}70.23, 73.22{]} & \cellcolor[HTML]{FB9273}67.24 \\
ROTS+max & \cellcolor[HTML]{F9746E}75.53 & {[}73.08, 77.77{]} & \cellcolor[HTML]{FCAA78}55.59 & {[}50.96, 59.70{]} & \cellcolor[HTML]{D8DF81}71.33 & {[}69.83, 72.80{]} & \cellcolor[HTML]{F96E6C}67.48 \\
ROTS+min & \cellcolor[HTML]{FCAF79}74.80 & {[}72.16, 77.14{]} & \cellcolor[HTML]{EAE582}53.44 & {[}48.65, 57.71{]} & \cellcolor[HTML]{DEE182}71.58 & {[}70.04, 73.00{]} & \cellcolor[HTML]{FDEA83}66.61 \\
ROTS+last & \cellcolor[HTML]{F9796E}75.47 & {[}72.99, 77.64{]} & \cellcolor[HTML]{FCA577}55.73 & {[}51.22, 59.76{]} & \cellcolor[HTML]{D3DE81}71.15 & {[}69.62, 72.62{]} & \cellcolor[HTML]{F9736D}67.45 \\\hline
with WR &  &  &  &  &  &  &  \\
WRD & \cellcolor[HTML]{DAE081}73.13 & {[}70.40, 75.54{]} & \cellcolor[HTML]{FA8270}56.73 & {[}52.07, 60.83{]} & \cellcolor[HTML]{90CB7D}68.73 & {[}67.13, 70.27{]} & \cellcolor[HTML]{D7DF81}66.20 \\
AC & \cellcolor[HTML]{63BE7B}70.13 & {[}67.35, 72.53{]} & \cellcolor[HTML]{89C97D}52.01 & {[}47.67, 56.15{]} & \cellcolor[HTML]{FDBC7B}73.20 & {[}71.66, 74.68{]} & \cellcolor[HTML]{74C37C}65.11 \\
ROTS+L0 & \cellcolor[HTML]{63BE7B}70.14 & {[}67.47, 72.65{]} & \cellcolor[HTML]{8BC97D}52.03 & {[}47.45, 56.02{]} & \cellcolor[HTML]{FDBC7B}73.20 & {[}71.63, 74.68{]} & \cellcolor[HTML]{75C37C}65.12 \\
ROTS+L1 & \cellcolor[HTML]{78C47C}70.67 & {[}67.98, 73.12{]} & \cellcolor[HTML]{91CB7D}52.13 & {[}47.62, 56.29{]} & \cellcolor[HTML]{FA7D6F}73.82 & {[}72.30, 75.28{]} & \cellcolor[HTML]{9BCE7E}65.54 \\
ROTS+L2 & \cellcolor[HTML]{8ECA7D}71.21 & {[}68.59, 73.62{]} & \cellcolor[HTML]{8BC97D}52.03 & {[}47.21, 56.24{]} & \cellcolor[HTML]{F8696B}74.01 & {[}72.52, 75.49{]} & \cellcolor[HTML]{AED37F}65.75 \\
ROTS+L3 & \cellcolor[HTML]{92CB7D}71.31 & {[}68.51, 73.77{]} & \cellcolor[HTML]{BCD780}52.76 & {[}48.20, 56.84{]} & \cellcolor[HTML]{F9716D}73.94 & {[}72.42, 75.40{]} & \cellcolor[HTML]{C5DA80}66.00 \\
ROTS+L4 & \cellcolor[HTML]{BBD780}72.34 & {[}69.78, 74.66{]} & \cellcolor[HTML]{E2E282}53.32 & {[}48.75, 57.45{]} & \cellcolor[HTML]{FA8471}73.75 & {[}72.16, 75.17{]} & \cellcolor[HTML]{F0E683}66.47 \\
ROTS+mean & \cellcolor[HTML]{9ACE7E}71.52 & {[}68.99, 73.87{]} & \cellcolor[HTML]{B1D47F}52.59 & {[}47.95, 56.78{]} & \cellcolor[HTML]{F97B6F}73.84 & {[}72.27, 75.28{]} & \cellcolor[HTML]{C4DA80}65.98 \\
ROTS+max & \cellcolor[HTML]{B4D57F}72.17 & {[}69.51, 74.51{]} & \cellcolor[HTML]{E2E282}53.32 & {[}48.78, 57.36{]} & \cellcolor[HTML]{FA8471}73.75 & {[}72.22, 75.20{]} & \cellcolor[HTML]{EBE582}66.41 \\
ROTS+min & \cellcolor[HTML]{63BE7B}70.12 & {[}67.20, 72.64{]} & \cellcolor[HTML]{63BE7B}51.44 & {[}46.76, 55.55{]} & \cellcolor[HTML]{FDBB7B}73.21 & {[}71.60, 74.66{]} & \cellcolor[HTML]{63BE7B}64.92 \\
ROTS+last & \cellcolor[HTML]{A4D17E}71.78 & {[}69.21, 74.10{]} & \cellcolor[HTML]{E8E482}53.41 & {[}49.05, 57.46{]} & \cellcolor[HTML]{FBA076}73.48 & {[}71.92, 74.96{]} & \cellcolor[HTML]{DAE081}66.22\\\hline
\end{tabular}
\end{table*}

\begin{table*}[]
\centering
\caption{Breakdown table with GloVe vectors}\label{app:tb:GloVe}
\scriptsize
\begin{tabular}{|l|p{1cm}l|p{1cm}l|p{1cm}l|p{1cm}|}
\hline
Similarity    & \multicolumn{2}{c|}{STSB}                           & \multicolumn{2}{c|}{Twitter}                        & \multicolumn{2}{c|}{SICK}                           & MEAN                          \\
          & Pearson's $r\times 100$                     & BCa 95\% CI        & Pearson's $r\times 100$                     & BCa 95\% CI        & Pearson's $r\times 100$                     & BCa 95\% CI        & Pearson's $r\times 100$                   \\\hline
with SUP       &                               &                    &                               &                    &                               &                    &                               \\
WRD       & \cellcolor[HTML]{FDC27D}71.97 & {[}69.18, 74.43{]} & \cellcolor[HTML]{FA8B72}55.63 & {[}50.93, 60.07{]} & \cellcolor[HTML]{89C97D}67.52 & {[}65.94, 69.13{]} & \cellcolor[HTML]{FFEA84}65.04 \\
AC       & \cellcolor[HTML]{DCE081}69.54 & {[}66.72, 72.05{]} & \cellcolor[HTML]{F3E783}49.79 & {[}44.88, 54.34{]} & \cellcolor[HTML]{FCA377}72.92 & {[}71.40, 74.43{]} & \cellcolor[HTML]{EAE582}64.08 \\
ROTS+L0   & \cellcolor[HTML]{DCE081}69.54 & {[}66.68, 72.13{]} & \cellcolor[HTML]{F3E783}49.79 & {[}44.83, 54.33{]} & \cellcolor[HTML]{FCA377}72.92 & {[}71.34, 74.40{]} & \cellcolor[HTML]{EAE582}64.08 \\
ROTS+L1   & \cellcolor[HTML]{E8E482}70.03 & {[}67.29, 72.51{]} & \cellcolor[HTML]{F7E883}50.1  & {[}45.13, 54.72{]} & \cellcolor[HTML]{F96E6C}73.51 & {[}71.95, 74.94{]} & \cellcolor[HTML]{F4E883}64.55 \\
ROTS+L2   & \cellcolor[HTML]{F7E883}70.62 & {[}67.89, 73.03{]} & \cellcolor[HTML]{F9E983}50.3  & {[}45.20, 54.90{]} & \cellcolor[HTML]{F8696B}73.56 & {[}72.03, 75.06{]} & \cellcolor[HTML]{FAE983}64.83 \\
ROTS+L3   & \cellcolor[HTML]{FFEB84}70.94 & {[}68.20, 73.40{]} & \cellcolor[HTML]{FFE583}51.08 & {[}46.17, 55.50{]} & \cellcolor[HTML]{FA7E6F}73.33 & {[}71.78, 74.74{]} & \cellcolor[HTML]{FFE583}65.12 \\
ROTS+L4   & \cellcolor[HTML]{FEC97E}71.79 & {[}69.19, 74.15{]} & \cellcolor[HTML]{FED680}51.83 & {[}46.91, 56.23{]} & \cellcolor[HTML]{FB9E76}72.98 & {[}71.44, 74.50{]} & \cellcolor[HTML]{FEC97E}65.53 \\
ROTS+mean & \cellcolor[HTML]{FFEB84}70.96 & {[}68.26, 73.37{]} & \cellcolor[HTML]{FFEB84}50.73 & {[}45.79, 55.23{]} & \cellcolor[HTML]{FA7C6F}73.35 & {[}71.79, 74.83{]} & \cellcolor[HTML]{FFEB84}65.01 \\
ROTS+max  & \cellcolor[HTML]{FED280}71.57 & {[}68.98, 73.95{]} & \cellcolor[HTML]{FED680}51.83 & {[}47.05, 56.38{]} & \cellcolor[HTML]{FB9E76}72.98 & {[}71.41, 74.43{]} & \cellcolor[HTML]{FECE7F}65.46 \\
ROTS+min  & \cellcolor[HTML]{E4E382}69.87 & {[}66.98, 72.43{]} & \cellcolor[HTML]{F1E683}49.54 & {[}44.59, 54.14{]} & \cellcolor[HTML]{FCA276}72.93 & {[}71.37, 74.40{]} & \cellcolor[HTML]{EBE582}64.11 \\
ROTS+last & \cellcolor[HTML]{FED580}71.49 & {[}68.90, 73.83{]} & \cellcolor[HTML]{FED781}51.76 & {[}47.01, 56.10{]} & \cellcolor[HTML]{FCAE79}72.8  & {[}71.21, 74.31{]} & \cellcolor[HTML]{FED580}65.35 \\\hline
with SWC       &                               &                    &                               &                    &                               &                    &                               \\
WRD       & \cellcolor[HTML]{FCB37A}72.34 & {[}69.61, 74.79{]} & \cellcolor[HTML]{F8696B}57.31 & {[}52.66, 61.52{]} & \cellcolor[HTML]{63BE7B}65.99 & {[}64.36, 67.51{]} & \cellcolor[HTML]{FFDE82}65.21 \\
AC       & \cellcolor[HTML]{FB9374}73.14 & {[}70.51, 75.51{]} & \cellcolor[HTML]{FB9073}55.34 & {[}50.75, 59.63{]} & \cellcolor[HTML]{E8E482}71.23 & {[}69.69, 72.67{]} & \cellcolor[HTML]{FA8370}66.57 \\
ROTS+L0   & \cellcolor[HTML]{FB9C75}72.93 & {[}70.19, 75.32{]} & \cellcolor[HTML]{FCA477}54.37 & {[}49.51, 58.72{]} & \cellcolor[HTML]{D6DF81}70.53 & {[}68.94, 72.02{]} & \cellcolor[HTML]{FCAD79}65.94 \\
ROTS+L1   & \cellcolor[HTML]{FB9173}73.19 & {[}70.55, 75.59{]} & \cellcolor[HTML]{FB9F76}54.6  & {[}49.87, 58.91{]} & \cellcolor[HTML]{E1E282}70.97 & {[}69.39, 72.39{]} & \cellcolor[HTML]{FB9875}66.25 \\
ROTS+L2   & \cellcolor[HTML]{FA8471}73.51 & {[}70.95, 75.90{]} & \cellcolor[HTML]{FB9B75}54.79 & {[}50.00, 59.19{]} & \cellcolor[HTML]{E0E282}70.94 & {[}69.41, 72.40{]} & \cellcolor[HTML]{FA8E72}66.41 \\
ROTS+L3   & \cellcolor[HTML]{FA7D6F}73.7  & {[}71.06, 76.06{]} & \cellcolor[HTML]{FA8A72}55.67 & {[}50.86, 59.99{]} & \cellcolor[HTML]{DAE081}70.68 & {[}69.15, 72.13{]} & \cellcolor[HTML]{F97C6F}66.68 \\
ROTS+L4   & \cellcolor[HTML]{F8696B}74.18 & {[}71.66, 76.50{]} & \cellcolor[HTML]{F97C6F}56.38 & {[}51.78, 60.67{]} & \cellcolor[HTML]{D0DD81}70.3  & {[}68.77, 71.77{]} & \cellcolor[HTML]{F8696B}66.95 \\
ROTS+mean & \cellcolor[HTML]{F97A6F}73.77 & {[}71.24, 76.14{]} & \cellcolor[HTML]{FB9173}55.29 & {[}50.58, 59.63{]} & \cellcolor[HTML]{DCE081}70.75 & {[}69.23, 72.22{]} & \cellcolor[HTML]{FA8170}66.60 \\
ROTS+max  & \cellcolor[HTML]{F9706D}74.01 & {[}71.44, 76.30{]} & \cellcolor[HTML]{F97C6F}56.38 & {[}51.71, 60.68{]} & \cellcolor[HTML]{D0DD81}70.3  & {[}68.73, 71.74{]} & \cellcolor[HTML]{F96D6C}66.90 \\
ROTS+min  & \cellcolor[HTML]{FB8F73}73.24 & {[}70.60, 75.63{]} & \cellcolor[HTML]{FCAB78}54.01 & {[}49.20, 58.35{]} & \cellcolor[HTML]{D6DF81}70.54 & {[}68.99, 72.06{]} & \cellcolor[HTML]{FCAE79}65.93 \\
ROTS+last & \cellcolor[HTML]{F9706D}74.01 & {[}71.43, 76.31{]} & \cellcolor[HTML]{F97A6F}56.45 & {[}51.79, 60.54{]} & \cellcolor[HTML]{CCDC81}70.13 & {[}68.60, 71.65{]} & \cellcolor[HTML]{F9706D}66.86 \\\hline
with WR        &                               &                    &                               &                    &                               &                    &                               \\
WRD       & \cellcolor[HTML]{CFDD81}69.05 & {[}65.99, 71.79{]} & \cellcolor[HTML]{E7E482}48.69 & {[}43.60, 53.49{]} & \cellcolor[HTML]{96CC7D}68.01 & {[}66.36, 69.60{]} & \cellcolor[HTML]{BBD780}61.92 \\
AC      & \cellcolor[HTML]{63BE7B}64.67 & {[}61.65, 67.48{]} & \cellcolor[HTML]{64BE7B}37.56 & {[}32.30, 42.53{]} & \cellcolor[HTML]{EBE582}71.36 & {[}69.72, 72.89{]} & \cellcolor[HTML]{63BE7B}57.86 \\
ROTS+L0   & \cellcolor[HTML]{63BE7B}64.67 & {[}61.65, 67.46{]} & \cellcolor[HTML]{64BE7B}37.56 & {[}32.23, 42.61{]} & \cellcolor[HTML]{EBE582}71.36 & {[}69.72, 72.91{]} & \cellcolor[HTML]{63BE7B}57.86 \\
ROTS+L1   & \cellcolor[HTML]{74C37C}65.39 & {[}62.39, 68.02{]} & \cellcolor[HTML]{6CC07B}38.24 & {[}33.05, 43.22{]} & \cellcolor[HTML]{FFEB84}72.12 & {[}70.52, 73.60{]} & \cellcolor[HTML]{72C27B}58.58 \\
ROTS+L2   & \cellcolor[HTML]{8CC97D}66.32 & {[}63.51, 68.98{]} & \cellcolor[HTML]{76C37C}39.12 & {[}33.70, 44.14{]} & \cellcolor[HTML]{FECC7E}72.47 & {[}70.85, 73.99{]} & \cellcolor[HTML]{82C77C}59.30 \\
ROTS+L3   & \cellcolor[HTML]{8FCA7D}66.46 & {[}63.59, 69.17{]} & \cellcolor[HTML]{87C87D}40.58 & {[}35.47, 45.37{]} & \cellcolor[HTML]{FDC27D}72.58 & {[}71.00, 74.06{]} & \cellcolor[HTML]{8ECA7D}59.87 \\
ROTS+L4   & \cellcolor[HTML]{B3D57F}67.9  & {[}65.24, 70.38{]} & \cellcolor[HTML]{95CC7D}41.77 & {[}36.62, 46.55{]} & \cellcolor[HTML]{FEC77E}72.52 & {[}70.93, 74.01{]} & \cellcolor[HTML]{A1D07E}60.73 \\
ROTS+mean & \cellcolor[HTML]{96CC7D}66.75 & {[}63.97, 69.35{]} & \cellcolor[HTML]{7CC57C}39.63 & {[}34.30, 44.41{]} & \cellcolor[HTML]{FED680}72.36 & {[}70.80, 73.91{]} & \cellcolor[HTML]{88C87D}59.58 \\
ROTS+max  & \cellcolor[HTML]{ADD37F}67.68 & {[}65.01, 70.09{]} & \cellcolor[HTML]{95CC7D}41.77 & {[}36.67, 46.65{]} & \cellcolor[HTML]{FEC77E}72.52 & {[}70.93, 74.02{]} & \cellcolor[HTML]{9FCF7E}60.66 \\
ROTS+min  & \cellcolor[HTML]{64BE7B}64.74 & {[}61.59, 67.57{]} & \cellcolor[HTML]{63BE7B}37.47 & {[}32.12, 42.42{]} & \cellcolor[HTML]{EBE582}71.37 & {[}69.71, 72.91{]} & \cellcolor[HTML]{63BE7B}57.86 \\
ROTS+last & \cellcolor[HTML]{A5D17E}67.36 & {[}64.63, 69.84{]} & \cellcolor[HTML]{90CB7D}41.31 & {[}36.23, 46.06{]} & \cellcolor[HTML]{FFE283}72.23 & {[}70.66, 73.70{]} & \cellcolor[HTML]{98CD7E}60.30\\\hline
\end{tabular}
\end{table*}

\begin{table*}[]
\centering
\caption{Breakdown table with Word2Vec vectors}\label{app:tb:Word2Vec}
\scriptsize
\begin{tabular}{|l|p{1cm}l|p{1cm}l|p{1cm}l|p{1cm}|}
\hline
Similarity    & \multicolumn{2}{c|}{STSB}                           & \multicolumn{2}{c|}{Twitter}                        & \multicolumn{2}{c|}{SICK}                           & MEAN                          \\
          & Pearson's $r\times 100$                     & BCa 95\% CI        & Pearson's $r\times 100$                     & BCa 95\% CI        & Pearson's $r\times 100$                     & BCa 95\% CI        & Pearson's $r\times 100$                   \\\hline
with SUP &  &  &  &  &  &  &  \\
WRD & \cellcolor[HTML]{FCA677}70.77 & {[}67.85, 73.44{]} & \cellcolor[HTML]{FA8871}41.36 & {[}36.04, 46.45{]} & \cellcolor[HTML]{64BE7B}66.34 & {[}64.73, 67.88{]} & \cellcolor[HTML]{FDB67A}59.49 \\
AC & \cellcolor[HTML]{B5D57F}69 & {[}66.05, 71.73{]} & \cellcolor[HTML]{B0D47F}33.12 & {[}27.54, 38.18{]} & \cellcolor[HTML]{F8E983}70.96 & {[}69.29, 72.53{]} & \cellcolor[HTML]{8FCA7D}57.69 \\
ROTS+L0 & \cellcolor[HTML]{B5D57F}69 & {[}66.05, 71.68{]} & \cellcolor[HTML]{B0D47F}33.12 & {[}27.55, 38.19{]} & \cellcolor[HTML]{F8E983}70.96 & {[}69.31, 72.52{]} & \cellcolor[HTML]{8FCA7D}57.69 \\
ROTS+L1 & \cellcolor[HTML]{D2DE81}69.42 & {[}66.39, 72.09{]} & \cellcolor[HTML]{BFD880}33.37 & {[}27.90, 38.44{]} & \cellcolor[HTML]{FFDD82}71.38 & {[}69.76, 72.93{]} & \cellcolor[HTML]{B8D67F}58.06 \\
ROTS+L2 & \cellcolor[HTML]{F7E883}69.95 & {[}66.94, 72.47{]} & \cellcolor[HTML]{D9E081}33.8 & {[}28.39, 38.93{]} & \cellcolor[HTML]{FDC47D}71.74 & {[}70.13, 73.26{]} & \cellcolor[HTML]{EAE582}58.50 \\
ROTS+L3 & \cellcolor[HTML]{FFE483}70.13 & {[}67.15, 72.70{]} & \cellcolor[HTML]{FFEB84}34.42 & {[}28.79, 39.60{]} & \cellcolor[HTML]{FED580}71.5 & {[}69.86, 73.02{]} & \cellcolor[HTML]{FFEB84}58.68 \\
ROTS+L4 & \cellcolor[HTML]{FDB97B}70.57 & {[}67.73, 73.13{]} & \cellcolor[HTML]{FFE483}34.97 & {[}29.38, 40.20{]} & \cellcolor[HTML]{FFEB84}71.17 & {[}69.51, 72.65{]} & \cellcolor[HTML]{FFDD82}58.90 \\
ROTS+mean & \cellcolor[HTML]{FAE983}69.99 & {[}67.02, 72.60{]} & \cellcolor[HTML]{E4E382}33.99 & {[}28.39, 39.10{]} & \cellcolor[HTML]{FED780}71.47 & {[}69.88, 73.03{]} & \cellcolor[HTML]{E9E482}58.48 \\
ROTS+max & \cellcolor[HTML]{FDB97B}70.57 & {[}67.73, 73.13{]} & \cellcolor[HTML]{FFE483}34.97 & {[}29.53, 40.05{]} & \cellcolor[HTML]{FFEB84}71.17 & {[}69.58, 72.74{]} & \cellcolor[HTML]{FFDD82}58.90 \\
ROTS+min & \cellcolor[HTML]{AED37F}68.91 & {[}65.89, 71.57{]} & \cellcolor[HTML]{B3D57F}33.18 & {[}27.90, 38.37{]} & \cellcolor[HTML]{F1E783}70.75 & {[}69.07, 72.25{]} & \cellcolor[HTML]{86C87D}57.61 \\
ROTS+last & \cellcolor[HTML]{FFE082}70.17 & {[}67.38, 72.81{]} & \cellcolor[HTML]{FFEA84}34.49 & {[}28.84, 39.53{]} & \cellcolor[HTML]{F0E683}70.72 & {[}69.09, 72.28{]} & \cellcolor[HTML]{E6E382}58.46 \\\hline
with SWC &  &  &  &  &  &  &  \\
WRD & \cellcolor[HTML]{FCB37A}70.64 & {[}67.76, 73.22{]} & \cellcolor[HTML]{F8696B}43.46 & {[}38.05, 48.49{]} & \cellcolor[HTML]{63BE7B}66.29 & {[}64.68, 67.83{]} & \cellcolor[HTML]{FA8C72}60.13 \\
AC & \cellcolor[HTML]{FEC97E}70.41 & {[}67.47, 73.05{]} & \cellcolor[HTML]{FCAC78}38.84 & {[}33.44, 44.03{]} & \cellcolor[HTML]{FFDC82}71.39 & {[}69.81, 72.89{]} & \cellcolor[HTML]{FA8771}60.21 \\
ROTS+L0 & \cellcolor[HTML]{FFE483}70.13 & {[}67.23, 72.81{]} & \cellcolor[HTML]{FCB179}38.46 & {[}33.08, 43.54{]} & \cellcolor[HTML]{F1E783}70.75 & {[}69.17, 72.26{]} & \cellcolor[HTML]{FCA377}59.78 \\
ROTS+L1 & \cellcolor[HTML]{FDC67D}70.44 & {[}67.59, 73.09{]} & \cellcolor[HTML]{FCAE79}38.67 & {[}33.18, 43.75{]} & \cellcolor[HTML]{FAE983}71.02 & {[}69.46, 72.51{]} & \cellcolor[HTML]{FB9273}60.04 \\
ROTS+L2 & \cellcolor[HTML]{FB9D75}70.87 & {[}68.02, 73.46{]} & \cellcolor[HTML]{FCA978}39.03 & {[}33.49, 43.93{]} & \cellcolor[HTML]{FFE984}71.2 & {[}69.61, 72.66{]} & \cellcolor[HTML]{FA7D6F}60.37 \\
ROTS+L3 & \cellcolor[HTML]{FA8D72}71.03 & {[}68.06, 73.57{]} & \cellcolor[HTML]{FCA176}39.58 & {[}33.98, 44.54{]} & \cellcolor[HTML]{F7E883}70.92 & {[}69.34, 72.40{]} & \cellcolor[HTML]{F9746D}60.51 \\
ROTS+L4 & \cellcolor[HTML]{F8696B}71.4 & {[}68.58, 73.98{]} & \cellcolor[HTML]{FB9B75}40.03 & {[}34.62, 45.28{]} & \cellcolor[HTML]{EBE582}70.56 & {[}69.02, 72.01{]} & \cellcolor[HTML]{F8696B}60.66 \\
ROTS+mean & \cellcolor[HTML]{FB9874}70.92 & {[}68.06, 73.46{]} & \cellcolor[HTML]{FCA777}39.21 & {[}33.82, 44.34{]} & \cellcolor[HTML]{F8E983}70.97 & {[}69.42, 72.43{]} & \cellcolor[HTML]{FA7D6F}60.37 \\
ROTS+max & \cellcolor[HTML]{F8696B}71.4 & {[}68.60, 73.96{]} & \cellcolor[HTML]{FB9B75}40.03 & {[}34.60, 45.49{]} & \cellcolor[HTML]{EBE582}70.56 & {[}68.98, 72.03{]} & \cellcolor[HTML]{F8696B}60.66 \\
ROTS+min & \cellcolor[HTML]{FFEB84}70.05 & {[}67.08, 72.66{]} & \cellcolor[HTML]{FCB179}38.51 & {[}33.04, 43.56{]} & \cellcolor[HTML]{ECE582}70.59 & {[}69.06, 72.13{]} & \cellcolor[HTML]{FCA777}59.72 \\
ROTS+last & \cellcolor[HTML]{F97B6F}71.22 & {[}68.32, 73.74{]} & \cellcolor[HTML]{FB9F76}39.74 & {[}34.26, 44.76{]} & \cellcolor[HTML]{E0E282}70.23 & {[}68.61, 71.71{]} & \cellcolor[HTML]{F97B6F}60.40 \\\hline
with WR &  &  &  &  &  &  &  \\
WRD & \cellcolor[HTML]{FED17F}70.33 & {[}67.42, 73.02{]} & \cellcolor[HTML]{FB9374}40.56 & {[}35.15, 45.83{]} & \cellcolor[HTML]{8FCA7D}67.68 & {[}66.09, 69.20{]} & \cellcolor[HTML]{FDB47A}59.52 \\
AC & \cellcolor[HTML]{65BE7B}67.86 & {[}64.83, 70.63{]} & \cellcolor[HTML]{63BE7B}31.85 & {[}26.34, 36.76{]} & \cellcolor[HTML]{FB9B75}72.33 & {[}70.74, 73.79{]} & \cellcolor[HTML]{68BF7B}57.35 \\
ROTS+L0 & \cellcolor[HTML]{65BE7B}67.87 & {[}64.88, 70.59{]} & \cellcolor[HTML]{63BE7B}31.86 & {[}26.51, 36.88{]} & \cellcolor[HTML]{FB9B75}72.33 & {[}70.74, 73.83{]} & \cellcolor[HTML]{69BF7B}57.35 \\
ROTS+L1 & \cellcolor[HTML]{8BC97D}68.4 & {[}65.42, 71.04{]} & \cellcolor[HTML]{70C17B}32.07 & {[}26.56, 37.15{]} & \cellcolor[HTML]{FA8370}72.68 & {[}71.08, 74.19{]} & \cellcolor[HTML]{92CB7D}57.72 \\
ROTS+L2 & \cellcolor[HTML]{BBD780}69.09 & {[}66.06, 71.70{]} & \cellcolor[HTML]{84C77C}32.4 & {[}26.71, 37.43{]} & \cellcolor[HTML]{F8696B}73.04 & {[}71.46, 74.49{]} & \cellcolor[HTML]{C6DA80}58.18 \\
ROTS+L3 & \cellcolor[HTML]{D4DE81}69.44 & {[}66.50, 72.09{]} & \cellcolor[HTML]{A6D17E}32.96 & {[}27.63, 38.06{]} & \cellcolor[HTML]{F9776E}72.85 & {[}71.33, 74.37{]} & \cellcolor[HTML]{E1E282}58.42 \\
ROTS+L4 & \cellcolor[HTML]{FBE983}70 & {[}67.15, 72.54{]} & \cellcolor[HTML]{C5DA80}33.48 & {[}27.87, 38.72{]} & \cellcolor[HTML]{FA8C72}72.55 & {[}70.94, 74.04{]} & \cellcolor[HTML]{FFEB84}58.68 \\
ROTS+mean & \cellcolor[HTML]{BFD880}69.14 & {[}66.29, 71.78{]} & \cellcolor[HTML]{90CB7D}32.6 & {[}27.20, 37.64{]} & \cellcolor[HTML]{F9796E}72.82 & {[}71.28, 74.29{]} & \cellcolor[HTML]{C7DB80}58.19 \\
ROTS+max & \cellcolor[HTML]{FBE983}70 & {[}67.12, 72.52{]} & \cellcolor[HTML]{C5DA80}33.48 & {[}27.99, 38.60{]} & \cellcolor[HTML]{FA8C72}72.55 & {[}70.99, 74.08{]} & \cellcolor[HTML]{FFEB84}58.68 \\
ROTS+min & \cellcolor[HTML]{63BE7B}67.83 & {[}64.78, 70.66{]} & \cellcolor[HTML]{68BF7B}31.94 & {[}26.48, 36.82{]} & \cellcolor[HTML]{FCA978}72.12 & {[}70.52, 73.67{]} & \cellcolor[HTML]{63BE7B}57.30 \\
ROTS+last & \cellcolor[HTML]{DDE182}69.58 & {[}66.77, 72.17{]} & \cellcolor[HTML]{AED37F}33.09 & {[}27.75, 38.21{]} & \cellcolor[HTML]{FCA376}72.22 & {[}70.62, 73.70{]} & \cellcolor[HTML]{D4DE81}58.30\\
\hline
\end{tabular}
\end{table*}

\begin{table*}[]
\centering
\caption{Breakdown table with PSL vectors}\label{app:tb:PSL}
\scriptsize
\begin{tabular}{|l|p{1cm}l|p{1cm}l|p{1cm}l|p{1cm}|}
\hline
Similarity    & \multicolumn{2}{c|}{STSB}                           & \multicolumn{2}{c|}{Twitter}                        & \multicolumn{2}{c|}{SICK}                           & MEAN                          \\
          & Pearson's $r\times 100$                     & BCa 95\% CI        & Pearson's $r\times 100$                     & BCa 95\% CI        & Pearson's $r\times 100$                     & BCa 95\% CI        & Pearson's $r\times 100$                   \\\hline
with SUP &  &  &  &  &  &  &  \\
WRD & \cellcolor[HTML]{FFEB84}73.78 & {[}71.22, 76.14{]} & \cellcolor[HTML]{F9786E}45.72 & {[}40.09, 51.01{]} & \cellcolor[HTML]{83C77C}67.83 & {[}66.23, 69.41{]} & \cellcolor[HTML]{E8E482}62.44 \\
AC & \cellcolor[HTML]{EEE683}73.50 & {[}70.84, 75.91{]} & \cellcolor[HTML]{D3DE81}42.49 & {[}36.81, 47.63{]} & \cellcolor[HTML]{FEEA83}71.97 & {[}70.38, 73.50{]} & \cellcolor[HTML]{FFEB84}62.65 \\
ROTS+L0 & \cellcolor[HTML]{EEE683}73.50 & {[}70.87, 75.82{]} & \cellcolor[HTML]{D3DE81}42.49 & {[}36.95, 47.70{]} & \cellcolor[HTML]{FFEB84}71.98 & {[}70.40, 73.47{]} & \cellcolor[HTML]{FFEB84}62.66 \\
ROTS+L1 & \cellcolor[HTML]{FDEA83}73.76 & {[}71.14, 76.07{]} & \cellcolor[HTML]{DEE182}42.71 & {[}37.24, 47.82{]} & \cellcolor[HTML]{FCA276}72.60 & {[}71.05, 74.07{]} & \cellcolor[HTML]{FCAD78}63.02 \\
ROTS+L2 & \cellcolor[HTML]{FECC7E}73.95 & {[}71.37, 76.21{]} & \cellcolor[HTML]{E3E282}42.81 & {[}37.08, 47.96{]} & \cellcolor[HTML]{FBA076}72.61 & {[}71.11, 74.08{]} & \cellcolor[HTML]{FB9C75}63.12 \\
ROTS+L3 & \cellcolor[HTML]{FECC7E}73.95 & {[}71.31, 76.32{]} & \cellcolor[HTML]{FFEB84}43.40 & {[}37.81, 48.61{]} & \cellcolor[HTML]{FDBD7C}72.37 & {[}70.84, 73.83{]} & \cellcolor[HTML]{FA8871}63.24 \\
ROTS+L4 & \cellcolor[HTML]{F8696B}74.48 & {[}71.98, 76.76{]} & \cellcolor[HTML]{FED981}43.78 & {[}38.22, 48.98{]} & \cellcolor[HTML]{FFE884}72.01 & {[}70.47, 73.45{]} & \cellcolor[HTML]{F8696B}63.42 \\
ROTS+mean & \cellcolor[HTML]{FB9F76}74.19 & {[}71.60, 76.44{]} & \cellcolor[HTML]{F0E683}43.10 & {[}37.48, 48.31{]} & \cellcolor[HTML]{FDBB7B}72.39 & {[}70.85, 73.86{]} & \cellcolor[HTML]{FA8B72}63.23 \\
ROTS+max & \cellcolor[HTML]{F9756E}74.42 & {[}71.87, 76.76{]} & \cellcolor[HTML]{FED981}43.78 & {[}38.04, 49.04{]} & \cellcolor[HTML]{FFE884}72.01 & {[}70.50, 73.48{]} & \cellcolor[HTML]{F96D6C}63.40 \\
ROTS+min & \cellcolor[HTML]{EFE683}73.52 & {[}70.90, 75.85{]} & \cellcolor[HTML]{CDDC81}42.36 & {[}36.69, 47.46{]} & \cellcolor[HTML]{FFEB84}71.98 & {[}70.40, 73.47{]} & \cellcolor[HTML]{FBE983}62.62 \\
ROTS+last & \cellcolor[HTML]{FB9674}74.24 & {[}71.63, 76.52{]} & \cellcolor[HTML]{FFDB81}43.74 & {[}38.23, 48.95{]} & \cellcolor[HTML]{F7E883}71.74 & {[}70.23, 73.25{]} & \cellcolor[HTML]{FA8871}63.24 \\\hline
with SWC &  &  &  &  &  &  &  \\
WRD & \cellcolor[HTML]{D1DD81}73.01 & {[}70.32, 75.35{]} & \cellcolor[HTML]{F8696B}46.01 & {[}40.33, 51.24{]} & \cellcolor[HTML]{63BE7B}66.73 & {[}65.19, 68.24{]} & \cellcolor[HTML]{AED37F}61.92 \\
AC & \cellcolor[HTML]{FB9A75}74.22 & {[}71.73, 76.54{]} & \cellcolor[HTML]{FFDA81}43.76 & {[}38.16, 48.85{]} & \cellcolor[HTML]{C6DA80}70.07 & {[}68.54, 71.57{]} & \cellcolor[HTML]{FFE684}62.68 \\
ROTS+L0 & \cellcolor[HTML]{FED07F}73.93 & {[}71.36, 76.25{]} & \cellcolor[HTML]{FFE182}43.62 & {[}38.07, 48.76{]} & \cellcolor[HTML]{B8D67F}69.60 & {[}68.00, 71.10{]} & \cellcolor[HTML]{E1E282}62.38 \\
ROTS+L1 & \cellcolor[HTML]{FDB87B}74.06 & {[}71.43, 76.41{]} & \cellcolor[HTML]{FED680}43.83 & {[}38.19, 49.15{]} & \cellcolor[HTML]{C8DB80}70.13 & {[}68.56, 71.66{]} & \cellcolor[HTML]{FFE884}62.67 \\
ROTS+L2 & \cellcolor[HTML]{FCA777}74.15 & {[}71.55, 76.46{]} & \cellcolor[HTML]{FED380}43.90 & {[}38.10, 49.26{]} & \cellcolor[HTML]{C9DB80}70.17 & {[}68.64, 71.64{]} & \cellcolor[HTML]{FFDD82}62.74 \\
ROTS+L3 & \cellcolor[HTML]{FCA978}74.14 & {[}71.58, 76.44{]} & \cellcolor[HTML]{FDB77A}44.45 & {[}38.75, 49.74{]} & \cellcolor[HTML]{C3D980}69.98 & {[}68.41, 71.45{]} & \cellcolor[HTML]{FEC97E}62.86 \\
ROTS+L4 & \cellcolor[HTML]{F96B6C}74.47 & {[}71.87, 76.78{]} & \cellcolor[HTML]{FCA577}44.82 & {[}39.21, 50.08{]} & \cellcolor[HTML]{BAD780}69.68 & {[}68.16, 71.12{]} & \cellcolor[HTML]{FCB37A}62.99 \\
ROTS+mean & \cellcolor[HTML]{FA7E6F}74.37 & {[}71.81, 76.73{]} & \cellcolor[HTML]{FDC47D}44.19 & {[}38.65, 49.45{]} & \cellcolor[HTML]{C3D980}69.98 & {[}68.41, 71.44{]} & \cellcolor[HTML]{FECB7E}62.85 \\
ROTS+max & \cellcolor[HTML]{F9776E}74.41 & {[}71.83, 76.67{]} & \cellcolor[HTML]{FCA477}44.83 & {[}39.20, 50.06{]} & \cellcolor[HTML]{BAD780}69.68 & {[}68.14, 71.13{]} & \cellcolor[HTML]{FDB57A}62.97 \\
ROTS+min & \cellcolor[HTML]{FDBD7C}74.03 & {[}71.44, 76.36{]} & \cellcolor[HTML]{FFE984}43.45 & {[}37.94, 48.60{]} & \cellcolor[HTML]{B8D67F}69.60 & {[}68.06, 71.15{]} & \cellcolor[HTML]{DEE182}62.36 \\
ROTS+last & \cellcolor[HTML]{FA7E6F}74.37 & {[}71.84, 76.66{]} & \cellcolor[HTML]{FCA276}44.88 & {[}39.25, 50.08{]} & \cellcolor[HTML]{B2D47F}69.41 & {[}67.83, 70.82{]} & \cellcolor[HTML]{FDC47D}62.89 \\\hline
with WR &  &  &  &  &  &  &  \\
WRD & \cellcolor[HTML]{B4D57F}72.52 & {[}69.80, 74.89{]} & \cellcolor[HTML]{FB9A75}45.04 & {[}39.41, 50.32{]} & \cellcolor[HTML]{94CC7D}68.38 & {[}66.77, 69.92{]} & \cellcolor[HTML]{B5D57F}61.98 \\
AC & \cellcolor[HTML]{63BE7B}71.13 & {[}68.34, 73.60{]} & \cellcolor[HTML]{66BE7B}40.18 & {[}34.60, 45.38{]} & \cellcolor[HTML]{FDBD7C}72.37 & {[}70.80, 73.85{]} & \cellcolor[HTML]{63BE7B}61.23 \\
ROTS+L0 & \cellcolor[HTML]{63BE7B}71.13 & {[}68.41, 73.61{]} & \cellcolor[HTML]{66BE7B}40.18 & {[}34.51, 45.23{]} & \cellcolor[HTML]{FDBD7C}72.37 & {[}70.77, 73.87{]} & \cellcolor[HTML]{63BE7B}61.23 \\
ROTS+L1 & \cellcolor[HTML]{7CC57C}71.57 & {[}68.95, 74.01{]} & \cellcolor[HTML]{73C27B}40.45 & {[}34.84, 45.47{]} & \cellcolor[HTML]{F96F6D}73.02 & {[}71.43, 74.48{]} & \cellcolor[HTML]{94CC7D}61.68 \\
ROTS+L2 & \cellcolor[HTML]{99CD7E}72.05 & {[}69.30, 74.44{]} & \cellcolor[HTML]{79C47C}40.59 & {[}34.86, 45.82{]} & \cellcolor[HTML]{F8696B}73.07 & {[}71.53, 74.54{]} & \cellcolor[HTML]{ACD37F}61.90 \\
ROTS+L3 & \cellcolor[HTML]{A3D07E}72.22 & {[}69.51, 74.62{]} & \cellcolor[HTML]{96CC7D}41.19 & {[}35.48, 46.47{]} & \cellcolor[HTML]{FA7E6F}72.90 & {[}71.36, 74.38{]} & \cellcolor[HTML]{C2D980}62.10 \\
ROTS+L4 & \cellcolor[HTML]{CCDC81}72.93 & {[}70.34, 75.21{]} & \cellcolor[HTML]{AAD27F}41.61 & {[}35.85, 46.86{]} & \cellcolor[HTML]{FCA276}72.60 & {[}71.00, 74.06{]} & \cellcolor[HTML]{E1E282}62.38 \\
ROTS+mean & \cellcolor[HTML]{AAD27F}72.34 & {[}69.75, 74.72{]} & \cellcolor[HTML]{87C87D}40.87 & {[}35.30, 46.18{]} & \cellcolor[HTML]{FA8070}72.88 & {[}71.31, 74.31{]} & \cellcolor[HTML]{BAD780}62.03 \\
ROTS+max & \cellcolor[HTML]{C4DA80}72.79 & {[}70.22, 75.05{]} & \cellcolor[HTML]{AAD27F}41.61 & {[}35.87, 46.84{]} & \cellcolor[HTML]{FCA276}72.60 & {[}71.04, 74.06{]} & \cellcolor[HTML]{DCE081}62.33 \\
ROTS+min & \cellcolor[HTML]{70C17B}71.36 & {[}68.54, 73.78{]} & \cellcolor[HTML]{63BE7B}40.11 & {[}34.50, 45.39{]} & \cellcolor[HTML]{FDBC7B}72.38 & {[}70.84, 73.91{]} & \cellcolor[HTML]{69BF7B}61.28 \\
ROTS+last & \cellcolor[HTML]{BBD780}72.63 & {[}70.04, 74.94{]} & \cellcolor[HTML]{A1D07E}41.43 & {[}36.01, 46.70{]} & \cellcolor[HTML]{FDBF7C}72.35 & {[}70.79, 73.85{]} & \cellcolor[HTML]{C6DA80}62.14\\\hline
\end{tabular}
\end{table*}

\begin{table*}[]
\centering
\caption{Breakdown table with ParaNMT vectors}\label{app:tb:ParaNMT}
\scriptsize
\begin{tabular}{|l|p{1cm}l|p{1cm}l|p{1cm}l|p{1cm}|}
\hline
Similarity    & \multicolumn{2}{c|}{STSB}                           & \multicolumn{2}{c|}{Twitter}                        & \multicolumn{2}{c|}{SICK}                           & MEAN                          \\
          & Pearson's $r\times 100$                     & BCa 95\% CI        & Pearson's $r\times 100$                     & BCa 95\% CI        & Pearson's $r\times 100$                     & BCa 95\% CI        & Pearson's $r\times 100$                   \\\hline
with SUP &  &  &  &  &  &  &  \\
WRD & \cellcolor[HTML]{CDDC81}79.05 & {[}76.85, 81.05{]} & \cellcolor[HTML]{F9706D}52.21 & {[}47.20, 56.89{]} & \cellcolor[HTML]{85C77C}70.02 & {[}68.52, 71.45{]} & \cellcolor[HTML]{FB9A75}67.09 \\
AC & \cellcolor[HTML]{FFEB84}79.55 & {[}77.23, 81.61{]} & \cellcolor[HTML]{FFEB84}46.56 & {[}41.08, 51.56{]} & \cellcolor[HTML]{FFEB84}73.89 & {[}72.47, 75.24{]} & \cellcolor[HTML]{FED280}66.67 \\
ROTS+L0 & \cellcolor[HTML]{FFEB84}79.55 & {[}77.15, 81.57{]} & \cellcolor[HTML]{FEEA83}46.55 & {[}41.14, 51.50{]} & \cellcolor[HTML]{FFEB84}73.89 & {[}72.45, 75.26{]} & \cellcolor[HTML]{FED380}66.66 \\
ROTS+L1 & \cellcolor[HTML]{FB9874}79.73 & {[}77.44, 81.77{]} & \cellcolor[HTML]{FFE683}46.81 & {[}41.44, 51.79{]} & \cellcolor[HTML]{FCB37A}74.47 & {[}73.09, 75.83{]} & \cellcolor[HTML]{FCA677}67.00 \\
ROTS+L2 & \cellcolor[HTML]{FB9874}79.73 & {[}77.40, 81.71{]} & \cellcolor[HTML]{FFE082}47.08 & {[}41.87, 52.25{]} & \cellcolor[HTML]{FCAA78}74.56 & {[}73.14, 75.88{]} & \cellcolor[HTML]{FB9674}67.12 \\
ROTS+L3 & \cellcolor[HTML]{FEEA83}79.54 & {[}77.09, 81.61{]} & \cellcolor[HTML]{FECA7E}48.09 & {[}42.85, 52.77{]} & \cellcolor[HTML]{FDC27D}74.31 & {[}72.91, 75.68{]} & \cellcolor[HTML]{FA7E6F}67.31 \\
ROTS+L4 & \cellcolor[HTML]{FB9374}79.74 & {[}77.41, 81.77{]} & \cellcolor[HTML]{FDBC7B}48.71 & {[}43.50, 53.50{]} & \cellcolor[HTML]{FFE884}73.93 & {[}72.55, 75.33{]} & \cellcolor[HTML]{F96A6C}67.46 \\
ROTS+mean & \cellcolor[HTML]{F9736D}79.81 & {[}77.48, 81.83{]} & \cellcolor[HTML]{FED580}47.60 & {[}42.36, 52.44{]} & \cellcolor[HTML]{FDC37D}74.30 & {[}72.90, 75.65{]} & \cellcolor[HTML]{FA8871}67.24 \\
ROTS+max & \cellcolor[HTML]{FB9374}79.74 & {[}77.43, 81.78{]} & \cellcolor[HTML]{FDBC7B}48.73 & {[}43.51, 53.49{]} & \cellcolor[HTML]{FFE884}73.93 & {[}72.51, 75.28{]} & \cellcolor[HTML]{F8696B}67.47 \\
ROTS+min & \cellcolor[HTML]{E5E382}79.29 & {[}76.79, 81.36{]} & \cellcolor[HTML]{EFE683}46.22 & {[}40.78, 51.23{]} & \cellcolor[HTML]{FFEB84}73.89 & {[}72.45, 75.25{]} & \cellcolor[HTML]{FDEA83}66.47 \\
ROTS+last & \cellcolor[HTML]{EDE582}79.37 & {[}77.02, 81.41{]} & \cellcolor[HTML]{FDBC7B}48.72 & {[}43.42, 53.61{]} & \cellcolor[HTML]{F6E883}73.63 & {[}72.16, 74.95{]} & \cellcolor[HTML]{FA8771}67.24 \\\hline
with SWC &  &  &  &  &  &  &  \\
WRD & \cellcolor[HTML]{63BE7B}77.98 & {[}75.65, 79.96{]} & \cellcolor[HTML]{F8696B}52.49 & {[}47.49, 57.00{]} & \cellcolor[HTML]{63BE7B}68.92 & {[}67.41, 70.35{]} & \cellcolor[HTML]{FDEA83}66.46 \\
AC & \cellcolor[HTML]{FCA677}79.70 & {[}77.58, 81.65{]} & \cellcolor[HTML]{FAE983}46.46 & {[}41.22, 51.39{]} & \cellcolor[HTML]{B6D67F}71.58 & {[}70.07, 72.98{]} & \cellcolor[HTML]{9CCE7E}65.91 \\
ROTS+L0 & \cellcolor[HTML]{FCA176}79.71 & {[}77.55, 81.57{]} & \cellcolor[HTML]{EEE683}46.20 & {[}40.78, 51.13{]} & \cellcolor[HTML]{ABD27F}71.23 & {[}69.75, 72.64{]} & \cellcolor[HTML]{78C47C}65.71 \\
ROTS+L1 & \cellcolor[HTML]{FA8170}79.78 & {[}77.56, 81.63{]} & \cellcolor[HTML]{FBE983}46.48 & {[}41.19, 51.48{]} & \cellcolor[HTML]{BCD780}71.77 & {[}70.30, 73.17{]} & \cellcolor[HTML]{ADD37F}66.01 \\
ROTS+L2 & \cellcolor[HTML]{FCA677}79.70 & {[}77.54, 81.64{]} & \cellcolor[HTML]{FFE583}46.84 & {[}41.49, 51.89{]} & \cellcolor[HTML]{BED880}71.85 & {[}70.40, 73.23{]} & \cellcolor[HTML]{C2D980}66.13 \\
ROTS+L3 & \cellcolor[HTML]{F7E883}79.47 & {[}77.14, 81.51{]} & \cellcolor[HTML]{FECD7F}47.97 & {[}42.69, 52.76{]} & \cellcolor[HTML]{B9D67F}71.67 & {[}70.27, 73.04{]} & \cellcolor[HTML]{ECE582}66.37 \\
ROTS+L4 & \cellcolor[HTML]{F8E883}79.48 & {[}77.34, 81.41{]} & \cellcolor[HTML]{FDBC7B}48.72 & {[}43.67, 53.62{]} & \cellcolor[HTML]{AFD47F}71.36 & {[}69.89, 72.76{]} & \cellcolor[HTML]{FFE583}66.52 \\
ROTS+mean & \cellcolor[HTML]{F97C6F}79.79 & {[}77.55, 81.74{]} & \cellcolor[HTML]{FFD981}47.40 & {[}42.14, 52.49{]} & \cellcolor[HTML]{B8D67F}71.64 & {[}70.14, 72.99{]} & \cellcolor[HTML]{DCE081}66.28 \\
ROTS+max & \cellcolor[HTML]{F5E883}79.45 & {[}77.24, 81.43{]} & \cellcolor[HTML]{FDBC7B}48.74 & {[}43.61, 53.64{]} & \cellcolor[HTML]{AFD47F}71.36 & {[}69.89, 72.74{]} & \cellcolor[HTML]{FFE683}66.52 \\
ROTS+min & \cellcolor[HTML]{FEC67D}79.63 & {[}77.33, 81.61{]} & \cellcolor[HTML]{E0E282}45.90 & {[}40.58, 50.83{]} & \cellcolor[HTML]{ABD37F}71.24 & {[}69.76, 72.68{]} & \cellcolor[HTML]{63BE7B}65.59 \\
ROTS+last & \cellcolor[HTML]{E4E382}79.28 & {[}77.07, 81.23{]} & \cellcolor[HTML]{FDBC7B}48.72 & {[}43.68, 53.51{]} & \cellcolor[HTML]{A5D17E}71.05 & {[}69.58, 72.47{]} & \cellcolor[HTML]{E9E482}66.35 \\\hline
with WR &  &  &  &  &  &  &  \\
WRD & \cellcolor[HTML]{CBDC81}79.03 & {[}76.80, 80.97{]} & \cellcolor[HTML]{FA8E73}50.82 & {[}45.66, 55.60{]} & \cellcolor[HTML]{A2D07E}70.94 & {[}69.47, 72.39{]} & \cellcolor[HTML]{FCB079}66.93 \\
AC & \cellcolor[HTML]{FDEA83}79.53 & {[}77.28, 81.52{]} & \cellcolor[HTML]{71C27B}43.46 & {[}38.01, 48.50{]} & \cellcolor[HTML]{FCAC78}74.54 & {[}73.07, 75.88{]} & \cellcolor[HTML]{8FCA7D}65.84 \\
ROTS+L0 & \cellcolor[HTML]{FDEA83}79.53 & {[}77.20, 81.46{]} & \cellcolor[HTML]{71C27B}43.46 & {[}38.11, 48.50{]} & \cellcolor[HTML]{FCAC78}74.54 & {[}73.08, 75.84{]} & \cellcolor[HTML]{8FCA7D}65.84 \\
ROTS+L1 & \cellcolor[HTML]{FA8F73}79.75 & {[}77.50, 81.75{]} & \cellcolor[HTML]{7DC57C}43.71 & {[}38.19, 48.46{]} & \cellcolor[HTML]{F9736D}75.12 & {[}73.70, 76.49{]} & \cellcolor[HTML]{CDDC81}66.19 \\
ROTS+L2 & \cellcolor[HTML]{FA8170}79.78 & {[}77.45, 81.75{]} & \cellcolor[HTML]{89C97D}43.98 & {[}38.40, 49.10{]} & \cellcolor[HTML]{F8696B}75.22 & {[}73.82, 76.56{]} & \cellcolor[HTML]{E5E382}66.33 \\
ROTS+L3 & \cellcolor[HTML]{FFEB84}79.55 & {[}77.15, 81.63{]} & \cellcolor[HTML]{BAD780}45.05 & {[}39.73, 49.92{]} & \cellcolor[HTML]{FA7D6F}75.02 & {[}73.61, 76.32{]} & \cellcolor[HTML]{FFE383}66.54 \\
ROTS+L4 & \cellcolor[HTML]{FCA176}79.71 & {[}77.41, 81.62{]} & \cellcolor[HTML]{D7DF81}45.70 & {[}40.42, 50.78{]} & \cellcolor[HTML]{FB9E76}74.68 & {[}73.29, 76.04{]} & \cellcolor[HTML]{FECE7F}66.70 \\
ROTS+mean & \cellcolor[HTML]{F8696B}79.83 & {[}77.57, 81.83{]} & \cellcolor[HTML]{A2D07E}44.54 & {[}39.07, 49.62{]} & \cellcolor[HTML]{FA7F70}75.00 & {[}73.62, 76.36{]} & \cellcolor[HTML]{FCEA83}66.46 \\
ROTS+max & \cellcolor[HTML]{FCA176}79.71 & {[}77.42, 81.64{]} & \cellcolor[HTML]{D8DF81}45.71 & {[}40.18, 50.60{]} & \cellcolor[HTML]{FB9E76}74.68 & {[}73.27, 76.01{]} & \cellcolor[HTML]{FECE7F}66.70 \\
ROTS+min & \cellcolor[HTML]{E4E382}79.28 & {[}76.85, 81.34{]} & \cellcolor[HTML]{63BE7B}43.14 & {[}37.45, 48.16{]} & \cellcolor[HTML]{FCAC78}74.54 & {[}73.10, 75.87{]} & \cellcolor[HTML]{6EC17B}65.65 \\
ROTS+last & \cellcolor[HTML]{EFE683}79.39 & {[}77.03, 81.36{]} & \cellcolor[HTML]{D4DE81}45.62 & {[}40.29, 50.65{]} & \cellcolor[HTML]{FDB97B}74.41 & {[}72.98, 75.77{]} & \cellcolor[HTML]{FFEB84}66.47
\\\hline
\end{tabular}
\end{table*}






\end{document}